%% file: root.tex
\documentclass[letterpaper, 10 pt, conference]{ieeeconf}  

\IEEEoverridecommandlockouts                            
\usepackage{math}
\usepackage{amsmath}

\let\labelindent\relax
\usepackage{enumitem}

\newcommand*\rot{\rotatebox{90}}

\title{\LARGE \bf
CSCPR: Cross-Source-Context Indoor RGB-D Place Recognition
\author{
Jing Liang$^{1}$, Zhuo Deng$^{2}$, Zheming Zhou$^{2}$, Min Sun$^{2}$, Omid Ghasemalizadeh$^{2}$, \\
Cheng-Hao Kuo$^{2}$, Arnie Sen$^{2}$, Dinesh Manocha$^{1}$
\thanks{$^{1}$ University of Maryland, College Park;  $^{2}$ Amazon, Bellevue, WA, USA }%
}
}

\newcommand{\rev}[1]{{\color{black}#1}}

\begin{document}

\maketitle
\thispagestyle{empty}
\pagestyle{empty}

\input{current_version}

\bibliographystyle{IEEEtran}
\bibliography{ref}

\newpage
\input{supplement}

\end{document}

%% file: current_version.tex
\noindent\begin{abstract}
\rev{We extend our previous work, PoCo~\cite{liang2024poco}, and present a new algorithm, Cross-Source-Context Place Recognition (CSCPR), for RGB-D indoor place recognition that integrates global retrieval and reranking into an end-to-end model and keeps the consistency of using Context-of-Clusters (CoCs)~\cite{cocs} for feature processing. Unlike prior approaches that primarily focus on the RGB domain for place recognition reranking, CSCPR is designed to handle the RGB-D data. We apply the CoCs to handle cross-sourced and cross-scaled RGB-D point clouds and introduce two novel modules for reranking: the Self-Context Cluster (SCC) and the Cross Source Context Cluster (CSCC), which enhance feature representation and match query-database pairs based on local features, respectively. We also release two new datasets, ScanNetIPR and ARKitIPR. Our experiments demonstrate that CSCPR significantly outperforms state-of-the-art models on these datasets by at least $29.27\%$ in Recall@1 on the ScanNet-PR dataset and $43.24\%$ in the new datasets. Website: \url{https://github.com/jingGM/PoCo-CCR.git}}

\end{abstract}


\section{Introduction}
\label{sec:intro}

Place recognition plays an important role in robotics~\cite{ijcai2021p603, lowry2015visual, liu2023survey, yurtsever2020survey}, where given query frames the goal is to identify matches from a database that share overlapping fields of view with queries based on image similarities~\cite{kornilova2023dominating, liu2023survey, yurtsever2020survey}. It is used in various applications, such as augmented reality~\cite{middelberg2014scalable}, navigation~\cite{mirowski2018learning, yurtsever2020survey}, SLAM~\cite{yurtsever2020survey, bresson2017simultaneous}, etc. However, the place recognition problem is very challenging~\cite{liu2023survey,yurtsever2020survey, lowry2015visual} due to: (1) different sensors, RGB, RGB-D, Lidar, etc., which require modality-specific feature processing; (2) environmental challenges, such as illumination changes, dynamic objects, occlusion, and scale variation; (3) the lack of study around RGB-D indoor place recognition~\cite{cgis, yurtsever2020survey, lowry2015visual}, where most of the current approaches only use global retrieval for place recognition~\cite{cgis, du2020dh3d}. Our approach focuses on the less explored domain of RGB-D indoor place recognition and proposes an end-to-end architecture with a reranking stage to handle the RGB-D place recognition task.

\begin{figure}
    \centering
    \includegraphics[width=\linewidth]{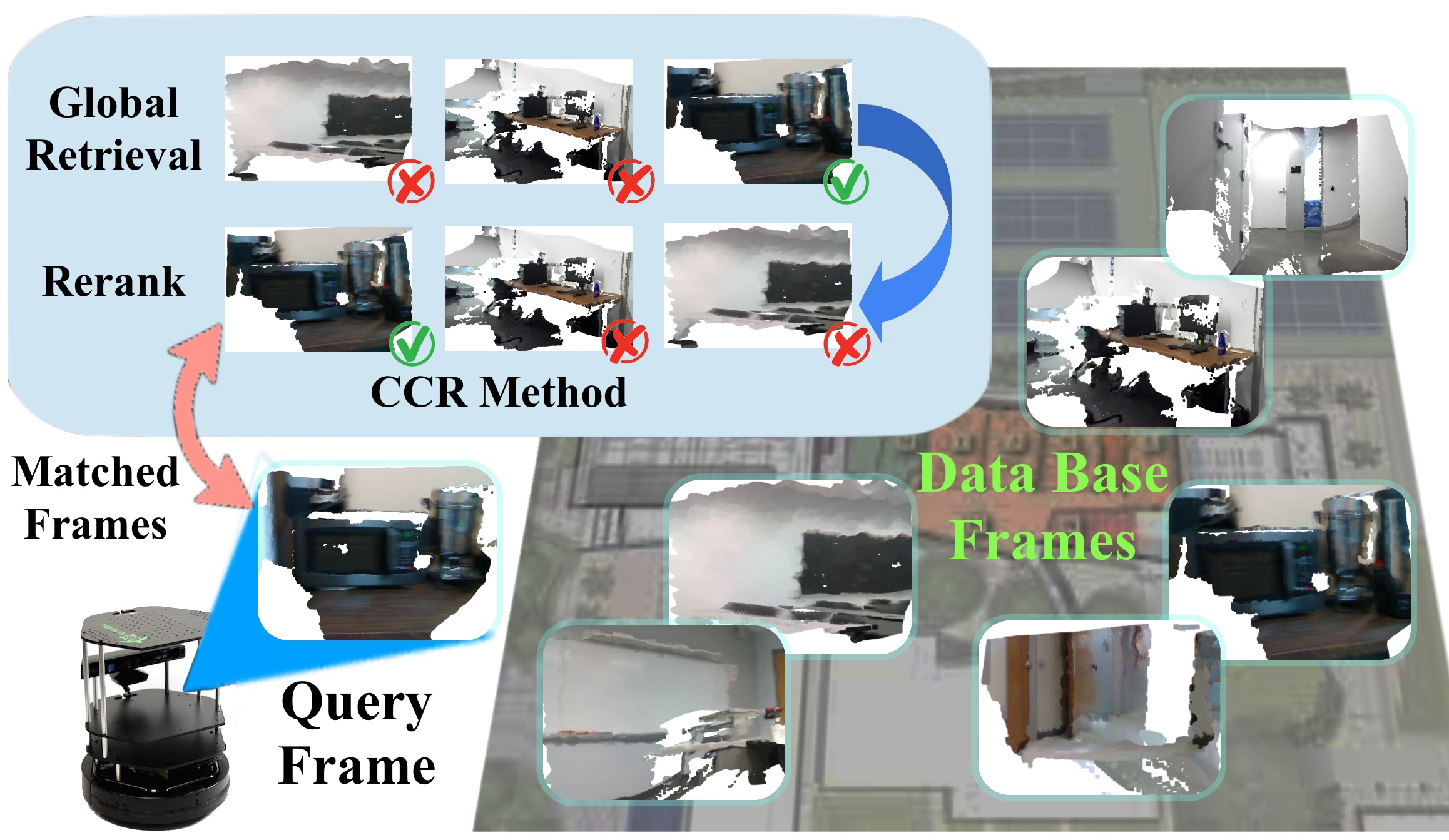}
    \vspace{-1.5em}
    \caption{\textbf{Real-world Experiment:} We propose a novel approach, Cross-Source-Context Place Recognition (CSCPR), for RGB-D indoor place recognition. Given a query frame, global retrieval ranks the potentially matched frames, and our novel place-recognition reranking model reranks the candidates to achieve better recognition accuracy.}
    \label{fig:front}
    \vspace{-2em}
\end{figure}

\textit{RGB-D indoor place recognition is not well studied:} 
Visual place recognition has been explored for many years, especially in the RGB domain~\cite{lowry2015visual, zhu2023r2former}. However, the potential of RGB-D data is overlooked; especially for indoor environments, the depth information can be crucial for place recognition~\cite{cgis, du2020dh3d, lowry2015visual}. For RGB-D place recognition, we need to handle different modalities of perceptions (color and geometry), so a good feature extractor is critical for the task. The CoCs~\cite{cocs} method shows comparable performance to the attention mechanism~\cite{vaswani2017attention}, and has been applied~\cite{liang2024poco} to handle different scales of features. Extending from our previous work, PoCo~\cite{liang2024poco}, we propose a novel end-to-end architecture, CSCPR, integrating global retrieval and reranking for RGB-D indoor place recognition, and we also keep the consistency of using the CoCs concept for feature processing.

\textit{RGB-D place-recognition reranking is not well studied:} 
Traditional learning-based RGB-D place recognition approaches~\cite{cgis, uy2018pointnetvlad, arandjelovic2016netvlad, yurtsever2020survey} rely on global retrieval, which extracts global descriptors from each frame and ranks the database frames by the similarities with the query frame. Reranking, as a complement to global retrieval, is meant to improve the accuracy by evaluating the matches of local features between the query and database frames~\cite{zhu2023r2former, hausler2021patch, vidanapathirana2023spectral}. Therefore, the reranking stage should be fast and accurate in matching local features. The RANSAC-based~\cite{ransac} methods are widely used for reranking by processing the geometric information of query-database frames~\cite{hausler2021patch, vidanapathirana2023spectral, tan2021instance}.
However, RANSAC is not a deep-learning-based approach and cannot be empirically trained to calculate the relationships among all local features. Recently, learning-based approaches~\cite{zhu2023r2former, lee2022correlation} have been proposed that match empirical local features. However, those approaches are all for RGB-only place recognition. In this work, we generalize the learning-based reranking concept to RGB-D point cloud and propose a novel learning-based method for RGB-D place-recognition reranking.

\textit{The scarcity of large-scale clean datasets is a notable gap for RGB-D indoor place recognition:} For indoor vision tasks, there are many datasets designed for object classification, segmentation, etc.~\cite{dai2017scannet, dehghan2021arkitscenes}.
However, those datasets are not specifically designed for training RGB-D place recognition tasks, which require positive and negative matched frames by overlap and a large scale for training purposes~\cite{kornilova2023dominating}. HPointLoc~\cite{yudin2022hpointloc} and ScanNet-PR~\cite{cgis} have recently been published for RGB-D indoor place recognition tasks, but their matched pairs are selected by the distance of camera poses or point cloud distance instead of the overlap of point clouds. This makes the dataset very noisy because of some nearby point clouds with no overlap, and it decreases the efficacy for training learning-based models. To bridge the gap, we introduce two large-scale clean datasets for RGB-D indoor place recognition by choosing frames with overlap.

\rev{\noindent \textbf{Main Results:} In this approach, we aim to solve the RGB-D indoor place recognition problem. We extend our previous work PoCo~\cite{liang2024poco} with a reranking stage and integrate the two stages into one end-to-end pipeline for place recognition. We keep the consistency of applying the CoCs concept for feature processing in the reranking stage and propose novel models to process multi-scale and multi-source features for RGB-D place recognition.
The following describes our contributions. Collectively, the contributions not only push the boundaries of RGB-D indoor place recognition to an integrated retrieval-reranking learning problem but also equip the community with a data generation pipeline and new datasets for the task.
\begin{enumerate}[leftmargin=*]
\item \textbf{Novel RGB-D Place-recognition Reranking:} We present a novel \textit{Self-Context Cluster (SCC)} to enhance local features by global information and a novel \textit{Cross Source Context Cluster (CSCC)} to process the local matches between different point clouds (sources) for fast and accurate reranking.
\item  \textbf{Curated Large-Scale RGB-D Indoor Place Recognition Datasets:} To address the scarcity of suitable large-scale RGB-D place-recognition datasets, we introduce ARKitIPR and ScanNetIPR datasets. We also propose a method to generate datasets by point-cloud overlap. The datasets contain positive and negative frames, keyframes for evaluation, poses, and semantic labels of point clouds.
\item \textbf{Superior Performance Across Multiple Datasets:} As shown in Tab.~\ref{tab:scannetpr_recall}, our approach outperforms other SOTA RGB-D indoor place recognition approaches by at least $29.27\%$ in ScanNet-PR~\cite{cgis} and $43.75\%$ in ScanNetIR and ARKitIPR. Compared with other place-recognition reranking methods, we improve at least $3.17$ points in the two novel datasets. We also demonstrate the effectiveness of our approach in a real-world robot.

\end{enumerate}
}

\section{Related Work}
\label{sec:related_work}

\noindent 
\textbf{Place Recognition Features and Methods:} Normally, place recognition is handled by calculating the similarities/overlap between the database and query frames. Traditional methods rely on RGB image features, such as SIFT, SURF, BoW,  etc.~\cite{david2004distinctive, cummins2008fab,galvez2012bags}, to compose descriptors for frame matching. After convolutional networks~\cite{gordo2017end} and attention mechanisms~\cite{vaswani2017attention} were proved to have better performance in vision tasks, various methods~\cite{gordo2017end, berton2022deep, touvron2021training} were proposed to encode frame features to global descriptors for place recognition, especially NetVLAD-based methods~\cite{uy2018pointnetvlad, arandjelovic2016netvlad}, which improve performance by combining convolutions with VLAD cores for global retrieval. The introduction of Context-of-Clusters (CoCs)~\cite{cocs} offers a comparably effective alternative but with different scales of feature processing, enhancing local features with global context. Our approach uses this concept to process and aggregate features of RGB-D point clouds.

For RGB-D place recognition tasks, NetVLAD~\cite{arandjelovic2016netvlad}, ORB~\cite{mur2017orb}, etc. use RGB information for place recognition by comparing the global descriptors of images. MinkLoc3D~\cite{komorowski2021minkloc3d} and Point-NetVLAD~\cite{uy2018pointnetvlad} process point clouds and compare point-cloud global descriptors. CGis-Net~\cite{cgis} applies KP-Conv~\cite{thomas2019kpconv} to encode the geometric information and enhance the features by extracting the semantic information from colors. However, these approaches typically only assess the global descriptors of RGB-D frames. We introduce a novel place-recognition reranking method to improve the accuracy of RGB-D indoor place recognition.

\noindent 
\textbf{Place-Recognition Reranking:} The accuracy of place recognition can be enhanced through a subsequent reranking stage following global retrieval~\cite{zhu2023r2former, hausler2021patch, zhang2022rank}. RANSAC-based geometric verification~\cite{vidanapathirana2023spectral, hausler2021patch} is a popular choice for reranking candidates. MAGSAC~\cite{Barath_2020_CVPR} uses a sample-based method NAPSAC to sample the points and apply a quality function to choose the matched pairs. PatchNet-VLAD~\cite{hausler2021patch} uses RANSAC with scoring methods for reranking after NetVLAD~\cite{arandjelovic2016netvlad}. Although RANSAC focuses on geometric verification, it overlooks non-geometric information. Similarly, registration tasks~\cite{urr, mei2023unsupervised, hatem2023point} also find inliers, but those methods still attempt to match points through geometric verification or rendering without empirically learning frame similarities. Matching methods~\cite{li20232d3d, sarlin2020superglue} like SuperGlue~\cite{sarlin2020superglue} identify inliers of the matching, but cannot guarantee if the frames are overlapped. Lee et al.~\cite{lee2022correlation} proposed CVNet-Rerank based on convolutional neural networks to process image features and compare the similarities among the features for reranking. R2Former~\cite{zhu2023r2former} used ViTs~\cite{dosovitskiy2020image} to extract the local information of images with higher attention values for reranking. However, those methods are all for RGB place recognition. RGB-D point cloud reranking still relies heavily on RANSAC-based geometric verification~\cite{vidanapathirana2023spectral}, or registration methods~\cite{zhang2022rank}. These approaches cannot use color and geometric information jointly. We propose an end-to-end structure to jointly process RGB and geometric information and we also integrate global retrieval and reranking together for RGB-D indoor place recognition.

\vspace{-0.5em}
\section{Method}
\label{sec:method}
In this section, we first formulate the problem in Section~\ref{sec:problem_formulation}. Then, in Section~\ref{sec:architecture} we describe the overall architecture of CSCPR, including the Self Context Cluster (SCC) and Cross-Source Context Cluster (CSCC) {modules}. Finally, the training strategies are described in Section~\ref{sec:training}.

\begin{figure*}
    \centering
\includegraphics[width=\linewidth]{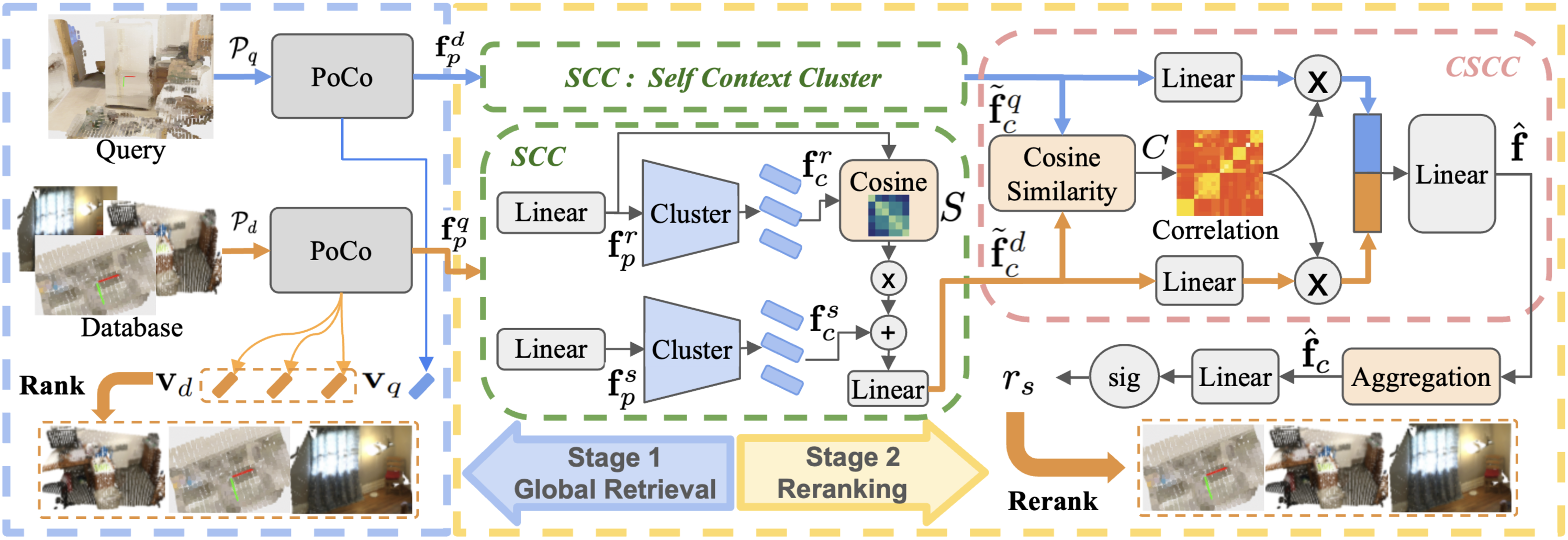}
    \vspace{-1.5em}
    \caption{\textbf{Architecture of CSCPR:} The blue box indicates Global Retrieval, and the yellow box represents Reranking. We use PoCo~\cite{liang2024poco} for global retrieval. After the database frames are ranked by comparing the global descriptors, $\v_q$ and $\v_d$, the reranking stage calculates the similarity of local features between query and database frames and reranks the database frames. The reranking model composes two Self Context Clusters (SCC) to process multi-scale features of each frame and a Cross Source Context Cluster (CSCC) to calculate the similarity of the local features between these two sorts of frames.
    }
    \label{fig:architecture}
    \vspace{-2em}
\end{figure*}
\subsection{Problem Definition}
\label{sec:problem_formulation}

As with other place recognition problems~\cite{liu2023survey, yurtsever2020survey, zhu2023r2former, arandjelovic2016netvlad}, the RGB-D indoor place recognition problem is defined as a retrieval problem. It's separated into two consecutive tasks, Global Retrieval~\cite{hausler2021patch, uy2018pointnetvlad, arandjelovic2016netvlad, cgis} and Reranking~\cite{zhang2022rank, zhu2023r2former}. In the paper, we use “frame” as an RGB-D colorized point cloud. As shown in Fig. \ref{fig:architecture}, CSCPR is an end-to-end structure to process the color and geometric information of the frames jointly. \rev{The Global Retrieval (blue box) is from our previous work, PoCo~\cite{liang2024poco}, with the same input format. It calculates the global descriptors, $\v_q$ of a query frame $\cp_q\in\cq$ and $\v_d$ of each of the database frames $\cp_d \in \cd$. Then we compare the similarity between $\v_q$ and all $\v_d$ to rank the database frames. However, the Global Retrieval model only matches the global descriptors, but some of the local information could be missed. The Reranking stage (yellow box) reranks database frames based on the similarities of local features, $\f_p^q$ and $\f_p^d$, between $\cp_q$ and $\cp_d$. The learning objective is to ensure the similarity between the {($\Q$, $\D_p$)} is much bigger than the similarity between {($\Q$, $\D_n$)}, where $\D_p \in \cd$ and $\D_n \in \cd$ correspond to the positive and negative matches to $\Q$.}

\subsection{Architecture of CSCPR}
\label{sec:architecture}
Considering that reranking is the second stage and compares local features of frames, the structure should: 1. \textbf{effectively process the relation} of the local features from different frames; 2. be \textbf{fast} and light in parameters and computation. To address these criteria, as shown in Fig.~\ref{fig:architecture}, we 1. propose a novel Self Context Clusters (SCC) to process multi-scale features and a novel Cross Source Context Cluster (CSCC) model to process the relation between two frames. 2. design a simple structure (yellow box) with only three small but powerful modules (two SCC and one CSCC).

\textbf{Self Context Cluster (SCC)} performs two functions: 1. \textbf{Small memory usage:} It downsamples the point features to a smaller number of center features. 2. \textbf{Multi-scale feature processing:} Because indoor RGB-D data have large variations w.r.t. perspectives, scales, and similar structures for different rooms, SCC learns features from different scales: local fine-scale feature $\f_p\in \mathcal{R}^{N\times D_p}$ with denser points and coarse-scale feature $\f_c\in \mathcal{R}^{M\times D_c}$ with sparser center points, where $N>M$ are the numbers of points and centers. To make the features more representative, we enhance each of the center features with the global information (all point features) of the frame.

\rev{As shown in the green box of Fig.~\ref{fig:architecture}, inspired by the attention mechanism~\cite{vaswani2017attention}, we transform the point features by projecting them into two distinct branches in SCC, reference point feature $\f_p^r=\norm{l_r(\f_p)}_g$ and source point feature $\f_p^s=\norm{l_s(\f_p)}_g$, where $l_r$ and $l_s$ are {different} linear layers to separate the point features into reference and source features. $\norm{\cdot}_g$ represents Group Norm.
The geometric information of the points is used to downsample the points to a smaller number of centers by the Farthest Downsampling method, and we find the K-nearest points for each center and average neighbors' features to a center feature $\f_c$. Thus, we have reference and source center features as $\f_c^r=\underset{k}{\text{mean}}(\f_p^r)$ and $\f_c^s=\underset{k}{\text{mean}}(\f_p^s)$.}

The current center features only contain local information about the frame, and we need to enhance it with the frame's global information, which is introduced by all the point features. Thus, we enhance the center features by calculating similarities between each center and all point features and aggregating the most globally similar point features to the center. The global similarity is calculated as Eq. \ref{eq:scc_sim}:
\begin{align}
    S = \text{sig} (\alpha \cos{(\f_c^r, \f_p^r)} + \beta),
    \label{eq:scc_sim}
\end{align}
where $\alpha$ and $\beta$ are trainable parameters. $\text{sig}$ and $\cos{}$ are Sigmoid and Cosine Similarity functions. The similarity matrix is {$S \in \mathcal{R}^{M \times N}$}. We threshold the similarity matrix to $\hat{S}$ by only using the most similar center for each point and setting other values as 0. To learn the sparse but representative multi-scale feature $\Tilde{\f}_c$ for reranking, according to the similarity matrix $S$, we aggregate the point features, $\f_{p}^r$, of the reference branch to center features $\f_{c}^s$ of the source branch and normalize them by the similarities:
\begin{align}
    \Tilde{\f}_{c,i} = l_c\left(\frac{1}{1 + \sum_{j=1}^N \hat{S}_{i,j}}\left(\f_{c,i}^s + \sum_{j=1}^N \hat{S}_{i, j} * \f_{p,j}^r\right)\right)
    \label{eq:aggregation_scc}
\end{align}
$i, j$ are the indices of centers and points. $l_c(\cdot)$ indicates linear layers. As demonstrated in Table~\ref{tab:rerank_ablation}, the learned “fine-coarse” scale features $\Tilde{\f_c}$ are more effective and efficient than Attention~\cite{vaswani2017attention}, which only processes single-scale features.

\rev{\textbf{Cross Source Context Cluster (CSCC)} captures cross-source correlation between the query and database frames. As in the red box of Fig.~\ref{fig:architecture}, the query and database frames have sparse multi-modality and multiple-scale center features $\Tilde{\f}_c^q$ and $\Tilde{\f}_c^d$, output from SCC. To calculate the correlation of the two frames, we use the same calculation method as Eq.~\ref{eq:scc_sim} but the points are from the same scale, $C = \text{sig} (\alpha \cos{(\Tilde{\f}_c^q, \Tilde{\f}_c^d)} + \beta)$, where $C \in \crr^{M_q \times M_d}$ is the correlation matrix and $M_q$ and $M_d$ are the center numbers of query and the database frames. We choose the top $K=500$ center pairs according to the correlation matrix, and other correlation values are masked out as 0. To fuse the multi-sourced feature together for reranking, given the masked correlation, we use it as weights to concatenate the query and retrieved candidate features together, where each concatenated feature is $\hat{\f}_{i,j} = l_c( C_{i,j}[l_q(\Tilde{\f}_{c, i}^q), l_d(\Tilde{\f}_{c,j}^d)])$, where $l_c, l_q, l_d$ indicating different linear layers. Finally, the aggregation function in Fig.~\ref{fig:architecture} aggregates the most correlated 500 features to one reranking score, $r_s$:}
\begin{equation}
    \hat{\f}_c = \frac{1}{1+\mathbf{c}_n}\sum_{j=1}^{M_d} (\frac{1}{1+\mathbf{c}_m}\sum_{i=1}^{M_q} \hat{\f}_{i,j})
\end{equation}
where $\mathbf{c}_m=\sum_{i=1}^{M_q} C_{i,j}$ is the aggregation of the correlation matrix along the dimension of query features and $\mathbf{c}_n=\underset{i}{\text{mean}}(\sum_{j=1}^{M_d} C_{i,j})$ is the aggregation of the correlation matrix along the dimension of database features and average in the dimension of query features. Then we have the reranking score of the database frames and the query frame to rerank database frames: $r_s = \text{sig}(l_f(\hat{\f}_c))$ , where $l_f$ indicates linear layers. Our overall design is simple but effective in capturing relationships between two RGB-D clouds (see Table~\ref{tab:rerank_ablation}).

\subsection{Training}
\label{sec:training}

Training the end-to-end place recognition, CSCPR, requires multiple loss functions for Global Retrieval and Reranking. A cosine annealing scheduler~\cite{loshchilov2016sgdr} and Adam optimizer~\cite{kingma2014adam} are used to schedule the learning rates and the learning rate changes from $10^{-4}$ to $10^{-7}$. Given ranked candidates from global retrieval, in the reranking stage we need to refine the rank by candidates' local features, so distinguishing between query and hard negatives is crucial in this stage. Therefore, we apply hard negative mining~\cite{robinson2020contrastive} in training and jointly train the two stages together. 

\rev{For Global Retrieval, we use the same loss functions as~\cite{liang2024poco} and we denote it as $\cl_g$. For Reranking, we use the entropy loss function:}
\begin{equation}
\cl_r = -(y * \log (r_s) + (1-y) * \log(1-r_s))~.
    \label{eq:cross_entropy}
\end{equation}
where $y$ is the label of the matching frames. $y=1$ when the database frame matches the query frame, and $y=0$ when they are not matched. The total loss function is $\cl = \beta_g\cl_g + \beta_r \cl_r$, where $\beta_g$ and $\beta_r$ are hyperparameters, and we choose 1.0 in the training.

\section{Dataset Generation}
\label{sec:data_generation}
\vspace{-1em}
\begin{algorithm}
\caption{Overlap is the major metric in dataset generation. 
The database frames are chosen by the overlap of frames from the original datasets and the positive and negative matches are chosen by overlaps with query frames.
}
\begin{algorithmic}
\Require $N$ consecutive frames, 
\Require $F_l \gets \{ \P_0, p_0 \}$
\Require database = $\{F_l, \}$
\While{$n < N$}
\State $F_n \gets \{ \P_n, p_n \}$
\If{FrameOverlap($F_l$, $F_n$) $< T_c$ }
    \State database.add($F_n$)
    \State $F_l = F_n$
\EndIf
\EndWhile

\end{algorithmic}
\label{alg:choose_database}
\end{algorithm}

\begin{figure}[ht!]
  \vspace{-2em}
  \centering
  \begin{tabular}{ | c | c |}
    \hline
    \textbf{FPF in ScanNetPR} & \textbf{FPF in ScanNetIPR (Ours)} \\
    \hline

    \begin{minipage}{.42\linewidth} \includegraphics[width=\linewidth]{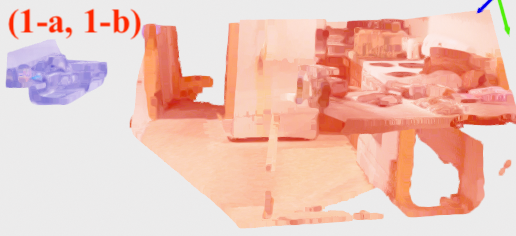} \end{minipage}
    &
    \begin{minipage}{.42\linewidth} \includegraphics[width=\linewidth]{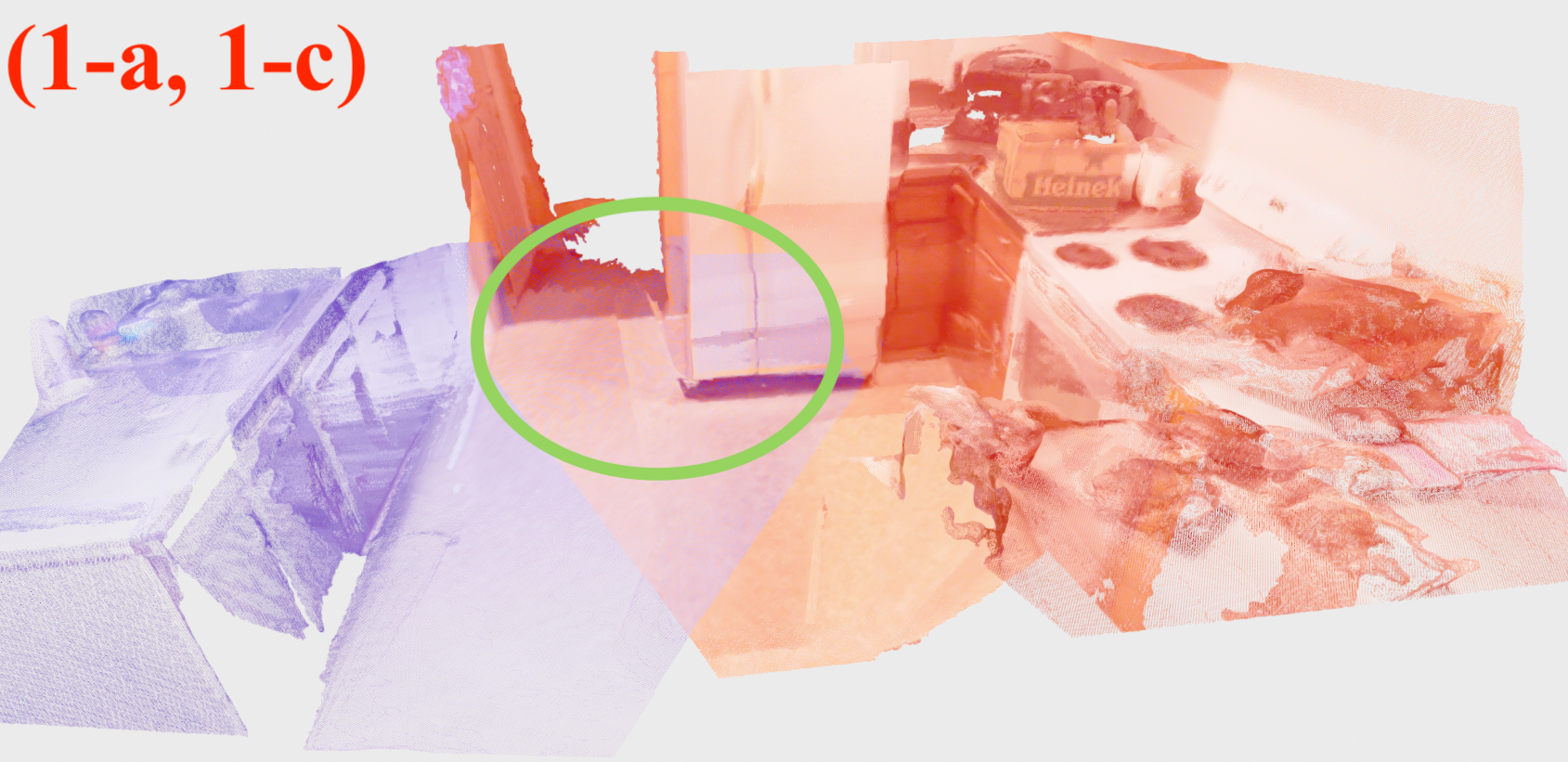} \end{minipage}
    \\ 
    \hline

  \end{tabular}
  \caption{\rev{\textbf{Furthest Positive Frame (FPF) of ScanNetPR vs. ScanNetIPR (ours):} FPF depicts the least overlapping matched frame to the query in both datasets. For the same query frame (red), ScanNetPR using center distance to determine matched frames leads to erroneous matching with no overlapped areas. In ScanNetIPR, the overlap (green) is the only criterion for matching; thus it is more accurate for training and evaluating place recognition task.}} 
  \label{fig:dataset_difference}
  \vspace{-1em}
\end{figure}

There are not many RGB-D datasets for indoor place recognition. ScanNetPR~\cite{cgis} tries to bridge the gap, but because the matching frames are chosen by the distance between frames' centers, there is some noise in the dataset. \rev{As shown in Fig.~\ref{fig:dataset_difference}, there are cases in which point clouds are very close, but they have no overlap. Without overlapping areas, the model cannot tell whether the two point clouds are in the same place, and it is also difficult to apply hard negative mining with the ScanNet-PR directly, limiting the final performance of the models.} To solve this issue and better evaluate the RGB-D indoor place recognition, we propose a dataset generation method and two large-scaled (in terms of the number of scenarios and frames) datasets, ScanNetIPR and ARKitIPR, based on ScanNet-V2~\cite{dai2017scannet} and ARKit~\cite{dehghan2021arkitscenes}. \rev{The generated ARKitIPR dataset has 5037 scenarios (4196 for training, 741 for evaluation, and 100 for testing) and 377625 frames of point clouds in total. ScanNetIPR has 1605 scenarios (1193 for training, 312 for evaluation, and 100 for testing) and 53201 frames of point clouds in total.} Both  datasets contain 1. positive and negative matching frames, 2. the pose of each frame, 3. database keyframes for the testing datasets and for recall calculation, and 4. point clouds that each contains positional $x,y,z$, normal $n_x,n_y,n_z$, color $r,g,b$ information and semantic labels. Because overlap of frames is the major reference for place recognition, we use overlap to determine positively and negatively matched cases for the dataset generation, as shown in Alg.~\ref{alg:choose_database}.

\textbf{First}, we generate database frames according to the coverage of frames from original datasets, ARKit and ScanNet-V2. In contrast to the odometry tasks, place recognition does not require very close frames to estimate the transformation matrices because those frames do not provide much different information for place retrieval and add burden in training. Therefore, the database frames are chosen by some overlap threshold along the trajectory of the camera motion, where if the current frame overlaps too much with the last frame it will be dropped, as shown in Alg. \ref{alg:choose_database}. The overlap in this step is calculated by the symmetric intersection of the union (IoU) of the voxelized frames. The FrameOverlap() in Alg.~\ref{alg:choose_database} is as in Eq.~\ref{eq:frame_overlap}, where $\P_n$ and $\P_l$ are points chosen by frustums in the scenario. $\v_n$ and $\v_l$ are the voxels of the $n_{th}$ frame and the last selected frame, respectively. $F_n = p_n \P_n$ represents the point cloud transformed in the global frame. The threshold $T_c = 0.5$ is used in the data generation task.
\begin{align}
    &[\v_n, \v_l] = 
    \text{voxelize}([p_n \P_n, \;p_l \P_l]), \;\;\; \\
    &\text{FrameOverlap}{(\P_n, \P_l)} = \frac{\abs{v_n \bigcap v_l}}{\abs{v_n \bigcup v_l}}.
    \label{eq:frame_overlap}
\end{align}

\textbf{Second}, we choose the positive frames in the same scenario and negative cases in all scenarios. Since this step treats each frame as a query frame to calculate the overlap with other frames in the scenario, the overlap is mostly w.r.t. the query frame. Therefore, an asymmetric overlap metric can be used:
\begin{align}
    \text{Overlap}{(\P_q, \P_d)} = \frac{\abs{v_q \bigcap v_o}}{\abs{v_q}}, 
    \label{eq:pn_frames}
\end{align}
where $\P_q$ and $\P_d$ are query and database frames. $\v_q$ and $\v_d$ are the voxels of these two frames with the same voxelization as Eq.~\ref{eq:frame_overlap}. Then the positive frames are selected by the threshold $\text{Overlap} > T_p$. Negative frames can be selected by $\text{Overlap} <= T_n$. Normally, the negative threshold $T_n=0$.

\textbf{Third}, we extract the key frames of each scenario, where we build a graph for each scenario using the positively matched frames and choose the dominating nodes as the keyframes of the scenario as \cite{kornilova2023dominating}.

\begin{table*}
    \centering
    \begin{tabular}{c| c |c|ccc|ccc} 
   \multirow{2}{*}{Approaches} & \multirow{2}{*}{Data Type}& {Inference}&\multicolumn{3}{c|}{\textit{ARKitIPR}} &\multicolumn{3}{c}{\textit{ScanNetIPR}}  \\ 
    &  & Time (ms) &R@1 $\uparrow$ &  R@2 $\uparrow$ &  R@3 $\uparrow$& R@1 $\uparrow$ &  R@2 $\uparrow$ &  R@3 $\uparrow$\\ \hline 
    PoCo~\cite{liang2024poco}                            &RGB-D  & -     & 45.12  & 57.10 & 62.14 & 58.10 & 70.33 & 75.95 \\\hline 
    Kabsch~\cite{arun1987least}          &RGB-D  & 93    & 53.00  & 59.37 & 65.59 & 66.07 & 74.89 & 80.99 \\ \hline
    TEASER~\cite{yang2020teaser}         &RGB-D  & 107   & 56.03  & 62.35 & 66.75 & 69.40 & 80.94 & 82.82 \\ \hline
    R2Former~\cite{zhu2023r2former}      &RGB-D  & 5     & 71.94  & 76.95 & 79.24 & 80.05 & 82.09 & 85.02 \\ \hline
    MAGSAC~\cite{barath2020magsac}       &RGB    & 18    & 20.60  & 33.03 & 38.49 & 41.00 & 47.25 & 52.30 \\ \hline
    SuperGlue~\cite{sarlin2020superglue} &RGB    & 40    & 49.19  & 55.50 & 59.44 & 69.84 & 75.83 & 78.72 \\ \hline
    SelaVPR~\cite{lu2024towards}         &RGB    & 11.4    &  56.32 & 63.81 & 67.56 & 67.31  & 76.47 & 80.60 \\ \hline
    URR~\cite{urr}                       &RGB-D  & 90    & 31.86  & 40.03 & 46.57 & 46.50 & 56.10 & 62.28 \\ \hline 
    PointMBF~\cite{pointmbf}             &RGB-D  & 300   & 50.12  & 57.30 & 63.47 & 61.27 & 70.92 & 77.40\\ \hline 
    Ours                                 &RGB-D  & 4     & \textbf{75.13} & \textbf{80.24} & \textbf{82.33} & \textbf{83.22} & \textbf{87.76} & \textbf{89.30} \\
    \end{tabular}
    \vspace{-0.5em}
    \caption{\textbf{Quantitative Reranking Results:} PoCo~\cite{liang2024poco} is the global retrieval baseline. Our place-recognition reranking method outperforms other reranking approaches by at least \textbf{3.19} points in Recall@1.} 
    \label{tab:rerank_recall}
    \vspace{-1em}
\end{table*}

\begin{table*}
    \centering
    \begin{tabular}{c| c |c|ccc|ccc} 
   \multirow{2}{*}{Approaches} & \multirow{2}{*}{Data Type}& {Inference}&\multicolumn{3}{c|}{\textit{ARKitIPR}} &\multicolumn{3}{c}{\textit{ScanNetIPR}}  \\ 
    &  & Time (ms) &R@1 $\uparrow$ &  R@2 $\uparrow$ &  R@3 $\uparrow$& R@1 $\uparrow$ &  R@2 $\uparrow$ &  R@3 $\uparrow$\\ \hline 
    Attention~\cite{vaswani2017attention}&RGB-D  & 11    & 64.14  & 67.65 & 68.81 & 74.95 & 82.94 & 86.22 \\ \hline 
    Ours/SCC                             &RGB-D  & 3     & 65.56  & 68.39 & 69.11 & 76.52 & 80.03 & 81.03 \\ \hline
    Ours/Correlation                     &RGB-D  & 4     & 45.56  & 60.18 & 66.06 & 47.33 & 65.99 & 74.95 \\ \hline
    Ours                                 &RGB-D  & 4     & \textbf{75.13} & \textbf{80.24} & \textbf{82.33} & \textbf{83.22} & \textbf{87.76} & \textbf{89.30} \\
    \end{tabular}
    \vspace{-0.5em}
    \caption{\textbf{Reranking Ablation Study:} 
    Ours, Ours/SCC, and Ours/Correlation represent our complete reranking, without SCC and without the correlation matrix, respectively. Attention represents the method with a sequence of self-cross attention blocks~\cite{vaswani2017attention}. The table shows effectiveness of our components and the outperformance w.r.t. an attention-based alternative by at least $11\%$ in ARKitIPR and ScanNetIPR.} 
    \label{tab:rerank_ablation}
        \vspace{-1em}
\end{table*}

\section{Experiment}
\label{sec:experiment}
\input{rerank_fig}

Our experimental design aims to evaluate the effectiveness and efficiency of our CSCPR against SOTA methods in RGB-D place recognition. To ensure a fair comparison, we conducted training and evaluation across multiple datasets, including ScanNet-PR, which uses a 3m threshold to select the database frames for place recognition, and newly proposed datasets, ScanNetIPR and ARKitIPR, which use overlap to select frames. For each dataset, we perform both training and testing for all models. The models are trained on 8 Tesla-V100 GPUs, and the input point clouds of CSCPR are constrained to 3000 points by voxelization downsampling. Evaluations are conducted in the device with an NVIDIA RTX A5000 GPU and an Intel Xeon(R) W-2255 CPU. The primary metric used for evaluation is the  Recall@1-3, which is the percentage of cases where at least one within top-k candidates is positive. Our experiments are structured into three distinct evaluations:

\noindent\textbf{E1: Place-Recognition Reranking Evaluation} This experiment is to evaluate the performance of our innovations, SCC and CSCC, on reranking. To make a fair comparison, we compare all the place-recognition reranking approaches with the same global retrieval stage.  We quantitatively and qualitatively demonstrate CSCPR's superior reranking performance in terms of accuracy and processing speed, which are our design criteria, over various SOTA approaches. 

\noindent\textbf{Comparisons with Other Reranking Approaches: } \textit{\textbf{a.} classical RANSAC-based geometric verification}, including RGB-based MAGSAC~\cite{barath2020magsac} to fit the homography of images~\cite{hausler2021patch} and RGB-D-based RANSAC + Kabsch~\cite{arun1987least}; \textit{\textbf{b.} learning-based place recognition reranking approaches.} Because of the lack of learning-based place recognition reranking works in the RGB-D domain, we adapt the SOTA method, R2Former~\cite{zhu2023r2former}, from the RGB domain to RGB-D point space by extending the pixel-position encoding method to $\{x,y,z\}$ positions. \textit{\textbf{c.} matching methods}, including RGB-based SuperGlue~\cite{sarlin2020superglue} and RGB-D-based URR~\cite{urr} to calculate the matching pairs and use the number of matched pairs to determine if the frames overlap. \textit{\textbf{d.} registration methods}~\cite{pointmbf, yang2020teaser, urr}, including TEASER++~\cite{yang2020teaser}, URR~\cite{urr}, and PointMBF~\cite{pointmbf}, which are mostly used for localization, where given a pair of matched frames, the corresponding features are extracted from overlapped areas and used to calculate transformation matrices. 

As shown in Tab~\ref{tab:rerank_recall}, we observe RGB-D-based approaches outperform corresponding RGB-based approaches. Our approach outperforms all other approaches by at least $3.19$ and $3.17$ at Recall@1 in ARKitIPR and ScanNetIPR, respectively. The learning-based approaches outperform RANSAC-based approaches. We also observe our approach has less inference time than other approaches by at least $20\%$, and this satisfies our design criterion; fast and effective. As shown in Fig~\ref{fig:qualitative_results}, there are two scenarios. We compare our approach, CSCPR, with the closest place-recognition reranking R2Former and TEASER++. R2Former, based on attention models, does not perform well in scenarios with very different scales but with similar geometric features, as shown in the first two rows of Fig.~\ref{fig:qualitative_results}. TEASER++ cannot match the color information of frames.

\noindent\textbf{Ablation Study:} As shown in Tab.~\ref{tab:rerank_ablation}, we compare our approach with three different modified versions to highlight the benefits of our design. Attention~\cite{vaswani2017attention} composes a sequence of self-attention and cross-attention models to substitute for our SCC and CSCC. The CoCs concept allows for multi-scale feature processing by relating large-scale environmental representations to small-scale object details, unlike attention mechanisms, which operate at a single scale without multi-scale interactions. Therefore, our design based on the CoCs concept is faster and outperforms attention-based alternatives by at least $11\%$ on ARKitIPR and ScanNetIPR in Recall@1. The attention model costs 12.5 Mb and 4.42 GMACs for a single batch, but our CSCPR costs 10 Mb and 3.01 GMACs, which is more computationally efficient. For the performance of single components, SCC improves the performance by at least $9\%$ on the two datasets by enhancing the local features with global information. We also observe the correlation matrix is critical in reranking and improves by at least $65\%$ on two datasets, as discussed in Section \ref{sec:architecture}, which provides the relationship between two frames. These results demonstrate the efficacy of our innovative components (SCC and CSCC) in RGB-D place recognition reranking.

\noindent\textbf{E2: End-to-End Solution Evaluation} As mentioned in Section~\ref{sec:intro}, RGB-D indoor place recognition reranking is not well explored, and we also did not find any integrated approaches with two stages for RGB-D place recognition. To make a fair comparison, we compare SOTA RGB-D approaches with global retrieval with CSCPR in ScanNet-PR~\cite{cgis}, as shown in Tab.~\ref{tab:scannetpr_recall}. Our approach outperforms other approaches by at least $29.27\%$ in Recall@1. 
This experiment also evaluates the performance of new RGB-D point cloud-based datasets, ScanNetIPR and ARKitIPR, as shown in Tab.~\ref{tab:new_data}. We observe at least $43.24\%$ improvement over other approaches in Recall@1 in the datasets.

\rev{\noindent\textbf{E3: Proposed Dataset Analysis} This experiment is to validate the performance of CSCPR but also to contribute valuable datasets to the community, facilitating further advancements in RGB-D indoor place recognition. Fig.~\ref{fig:dataset_difference} shows the benefits of our datasets with less noise in the datasets. 
From the comparison of Tab.~\ref{tab:scannetpr_recall} and \ref{tab:new_data}, we observe approaches in our ScanNetIPR dataset have lower recall, meaning our ScanNetIPR dataset is more difficult than ScanNet-PR. The reason is ScanNetPR has more positive frames, 19.94, on average for each frame in each scenario than ScanNetIPR, which has 15.98 on average. From Tab.~\ref{tab:new_data}, we show our new generated datasets are valid for RGB-D indoor place recognition with performance comparable to the results in Tab.~\ref{tab:scannetpr_recall}.
}
\begin{table}[!h]
\vspace{-0.5em}
\centering
\begin{tabular}{c|c|ccc}
\textbf{\textit{ScanNet-PR}} & Data Type & R@1 $\uparrow$ &  R@2 $\uparrow$ &  R@3 $\uparrow$\\
\hline
SIFT~\cite{david2004distinctive} + BoW~\cite{sivic2008efficient} & RGB &  16.16 & 21.17 & 24.38 \\
\hline
NetVLAD~\cite{arandjelovic2016netvlad} & RGB &21.77 & 33.81 & 41.49 \\
\hline
PointNetVLAD~\cite{uy2018pointnetvlad} & Point Cloud &27.10 & 32.10 & 37.01 \\
\hline
MinkLoc3D~\cite{komorowski2021minkloc3d} & Point Cloud &15.21 & 19.25 & 22.79 \\
\hline
Indoor DH3D~\cite{du2020dh3d} & RGB-D & 16.10 & 21.92 & 25.30 \\
\hline
CGiS-Net~\cite{cgis} &RGB-D & 61.12 & 70.23 & 75.06 \\
\hline
PoCo~\cite{liang2024poco} & RGB-D & 64.63 & 75.02 & 80.09 \\\hline
AEGIS-NET~\cite{ming2024aegis} & RGB-D & 65.09 & 74.26 & 79.06 \\
\hline
CGiS-Net w/o color~\cite{cgis} & Point Cloud & 39.62 & 50.92 & 56.14 \\
\hline
PoCo~\cite{liang2024poco} w/o color & Point Cloud & 44.34 & 54.27 & 59.78\\
\hline
\hline
CSCPR w/o color & Point Cloud & 60.58 & 73.59 & 78.29 \\
\hline
CSCPR & RGB-D & \textbf{84.14} & \textbf{ 89.82} & \textbf{91.25} \\
\end{tabular}
\caption{\rev{\textbf{Quantitative Results:} Compared with other SOTA methods, our approach CSCPR with reranking improved the Recall@1 performance by at least $\mathbf{29.27\%}$.}} 
\label{tab:scannetpr_recall}
    \vspace{-1.5em}
\end{table}

\begin{table}[!ht]
    \centering
\begin{tabular}{c|c|c|ccc}
 & Approaches  & Data Type   & R@1 $\uparrow$ & R@2 $\uparrow$ &  R@3 $\uparrow$\\ \hline 
\multirow{5}{*}{\rot{\textbf{\textit{ScanNetIPR}}}} & PointNetVLAD &Point Cloud& 22.43 & 30.81 & 36.58 \\ 
&MinkLoc3D &Point Cloud& 10.13 & 16.63 & 20.80 \\ 
&CGiS-Net  &RGB-D  & 57.89     & 69.95 & 75.51 \\  
&AEGIS-NET~\cite{ming2024aegis} & RGB-D & 58.00 & 69.12 & 74.79 \\
&PoCo~\cite{liang2024poco} &RGB-D            & 58.10          & 70.33          & 75.95 \\
&CSCPR                       &RGB-D      & \textbf{83.22} & \textbf{87.76} & \textbf{89.30} \\ \hline

\multirow{5}{*}{\rot{\textbf{\textit{ARKitIPR}}}} &PointNetVLAD &Point Cloud& 11.04 & 16.57 & 20.57 \\ 
&MinkLoc3D &Point Cloud& 8.14 & 10.95 & 13.79 \\ 
&CGiS-Net&RGB-D & 39.80 & 49.30 & 55.64 \\  
&AEGIS-NET~\cite{ming2024aegis} & RGB-D & 45.71 & 57.71 & 63.34 \\
&PoCo~\cite{liang2024poco} &RGB-D& 45.12 &  57.10 & 62.14 \\
&CSCPR &RGB-D& \textbf{75.13} & \textbf{80.24} & \textbf{82.33} \\

    \end{tabular}
    \caption{\rev{\textbf{Comparisons in New ScanNetIPR and ARKitIPR:} The ScanNetIPR is more difficult than ScanNetPR, where methods have relatively smaller Recall@1. Our method still outperforms other approaches for both RGB-D and pure point-cloud place recognition in both ScanNetIPR and ARKitIPR by at least $43.24\%$ in Recall@1.}} 
    \label{tab:new_data}
    \vspace{-2.5em}
\end{table}

\section{Conclusion, Limitations, and Future Work}
We explored RGB-D place recognition with an integrated pipeline with both global retrieval and reranking. By developing a fast and effective reranking model, we close the gap between RGB-D place recognition reranking and learning-based algorithms. We generalized the CoCs concept to noisy colorized point cloud feature processing and demonstrated better performance in place recognition tasks. We handle the scarcity of RGB-D place recognition datasets and propose a data generation pipeline for the community to explore more datasets. We introduce two large-scale RGB-D datasets for training and testing purposes. We push forward the boundary of RGB-D indoor place recognition accuracy by demonstrating that our design outperforms other SOTA approaches by at least $29.27\%$ in Recall@1 in the ScanNet-PR dataset. For limitations, if the overlapping areas do not have many features, our method may not work well. As part of future work, we would like to apply semantic pretraining to the model to improve its understanding of the environment.

%% file: rerank_fig.tex
\begin{figure*}[!ht]
  \centering
  \begin{tabular}{ | c | c | c | c | c |}
    \hline
   \textbf{Query Frames} & \textbf{PoCo~\cite{liang2024poco} } & \textbf{CSCPR } & \textbf{R2former } & \textbf{TEASER++}  \\ \hline

    \begin{minipage}{.17\linewidth} \includegraphics[width=\linewidth,height=0.75\linewidth]{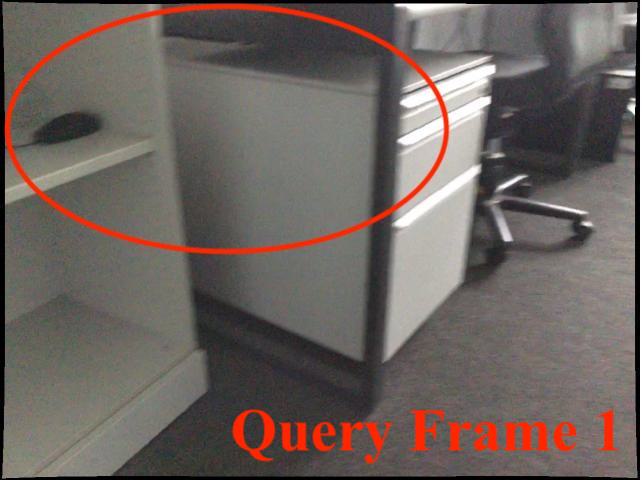} \end{minipage}
    &
    \begin{minipage}{.17\linewidth} \includegraphics[width=\linewidth,height=0.75\linewidth]{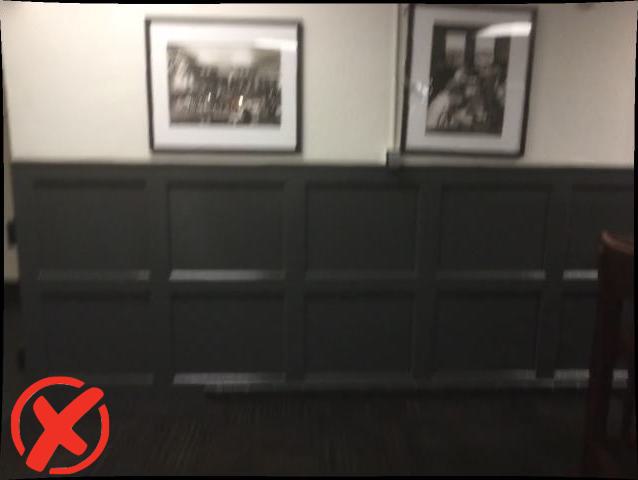} \end{minipage}
    &
    \begin{minipage}{.17\linewidth} \includegraphics[width=\linewidth,height=0.75\linewidth]{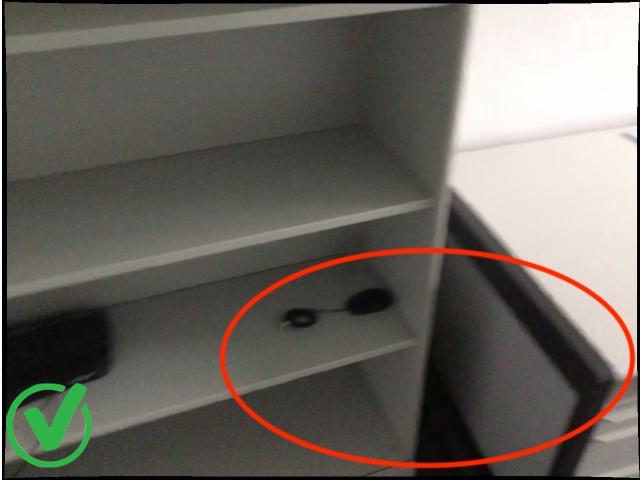} \end{minipage}
    &
    \begin{minipage}{.17\linewidth} \includegraphics[width=\linewidth,height=0.75\linewidth]{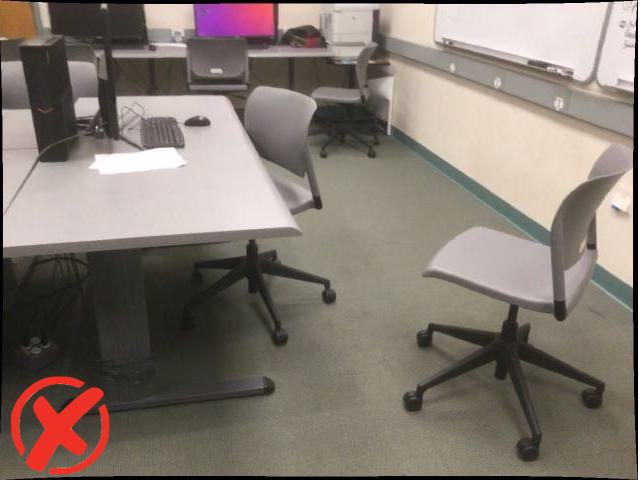} \end{minipage}
    &
    \begin{minipage}{.17\linewidth} \includegraphics[width=\linewidth,height=0.75\linewidth]{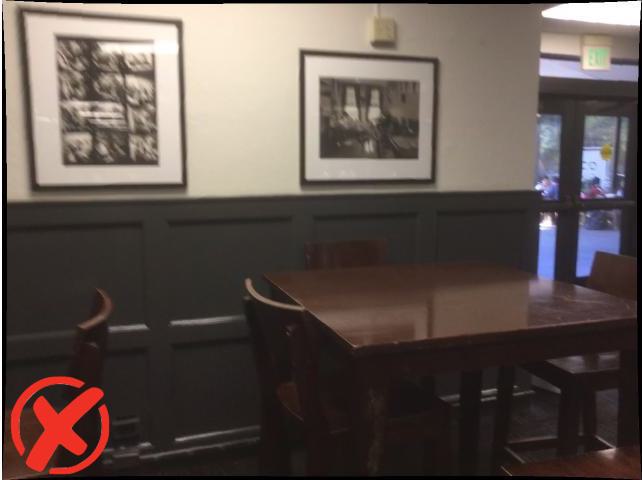} \end{minipage}
    \\ \hline
    \begin{minipage}{.17\linewidth} \includegraphics[width=\linewidth,height=0.75\linewidth]{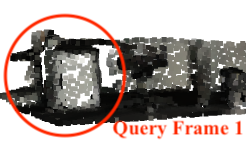} \end{minipage}
    &
    \begin{minipage}{.17\linewidth} \includegraphics[width=\linewidth,height=0.75\linewidth]{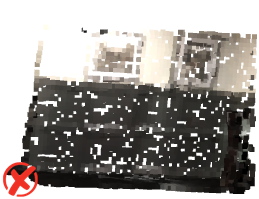} \end{minipage}
    &
    \begin{minipage}{.17\linewidth} \includegraphics[width=\linewidth,height=0.75\linewidth]{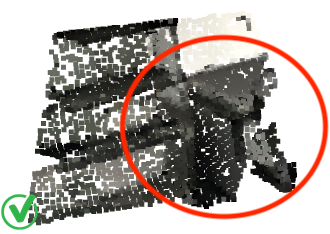} \end{minipage}
    &
    \begin{minipage}{.17\linewidth} \includegraphics[width=\linewidth,height=0.75\linewidth]{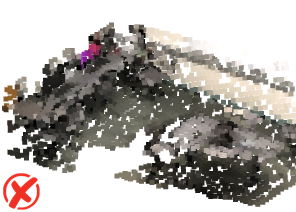} \end{minipage}
    &
    \begin{minipage}{.17\linewidth} \includegraphics[width=\linewidth,height=0.75\linewidth]{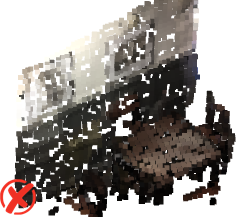} \end{minipage}
    \\ \hline\hline
    \begin{minipage}{.17\linewidth} \includegraphics[width=\linewidth,height=0.75\linewidth]{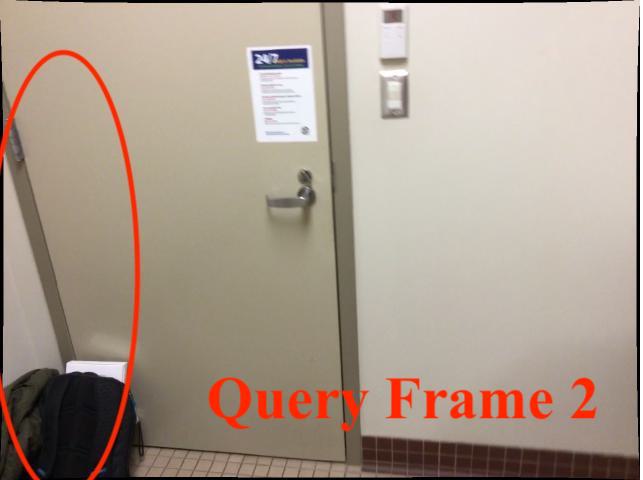} \end{minipage}
    &
    \begin{minipage}{.17\linewidth} \includegraphics[width=\linewidth,height=0.75\linewidth]{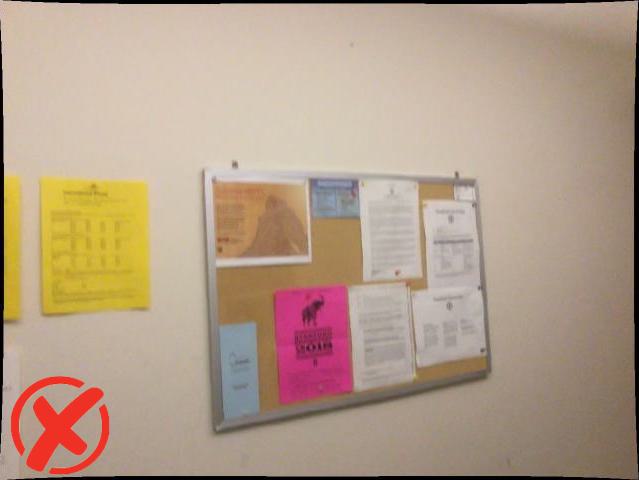} \end{minipage}
    &
    \begin{minipage}{.17\linewidth} \includegraphics[width=\linewidth,height=0.75\linewidth]{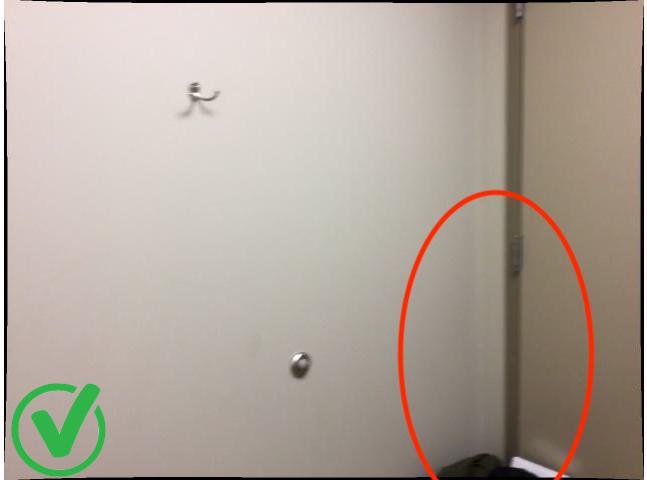} \end{minipage}
    &
    \begin{minipage}{.17\linewidth} \includegraphics[width=\linewidth,height=0.75\linewidth]{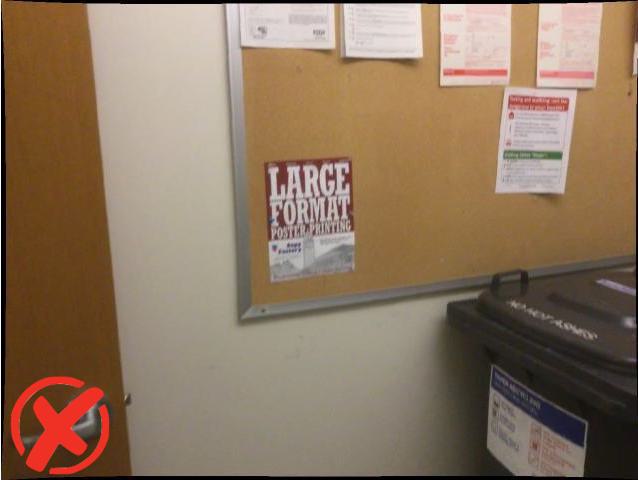} \end{minipage}
    &
    \begin{minipage}{.17\linewidth} \includegraphics[width=\linewidth,height=0.75\linewidth]{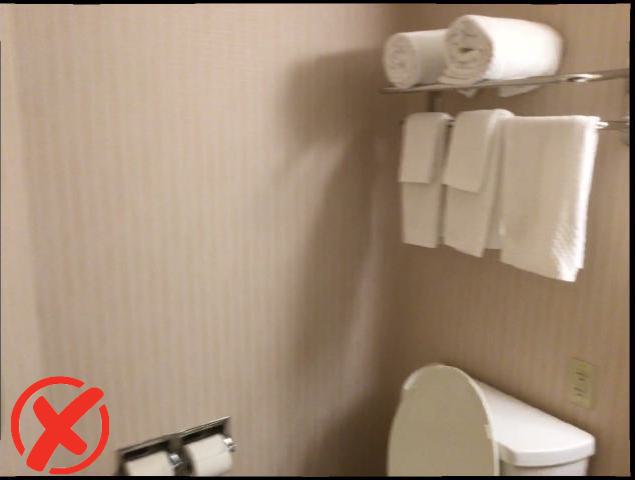} \end{minipage}
    \\ \hline
    
    \begin{minipage}{.17\linewidth} \includegraphics[width=\linewidth,height=0.75\linewidth]{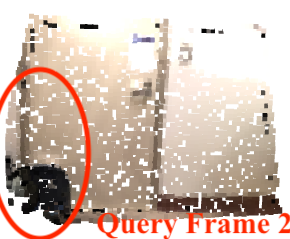} \end{minipage}
    &
    \begin{minipage}{.17\linewidth} \includegraphics[width=\linewidth,height=0.75\linewidth]{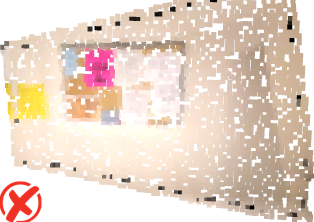} \end{minipage}
    &
    \begin{minipage}{.17\linewidth} \includegraphics[width=\linewidth,height=0.75\linewidth]{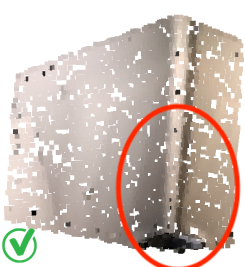} \end{minipage}
    &
    \begin{minipage}{.17\linewidth} \includegraphics[width=\linewidth,height=0.75\linewidth]{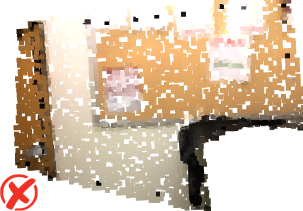} \end{minipage}
    &
    \begin{minipage}{.17\linewidth} \includegraphics[width=\linewidth,height=0.75\linewidth]{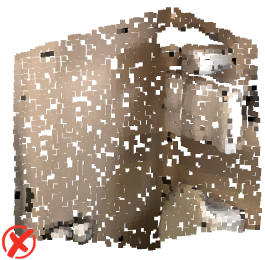} \end{minipage}
    \\ \hline
  \end{tabular}
  \caption{\rev{\textbf{Qualitative Comparisons: } 1st and 3rd rows show RGB images corresponding to point clouds in 2nd and 4th rows. The red circles mark the overlapping areas between query frames (1st column) and later Recall@1 frames from different approaches. R2Former performs closest to our approach, but it does not perform well in different scaled frames. Our overall algorithm (CSCPR) balances the geometric and RGB information well and achieves the best performance, even for the scenarios that have very small overlapping areas.}}
  \label{fig:qualitative_results}
  \vspace{-2em}
\end{figure*}

%% file: supplement.tex
\clearpage

\section{Supplement}
\label{sec:supplement}

\begin{figure}[ht!]
  \centering
    \noindent\includegraphics[width=0.9\linewidth]{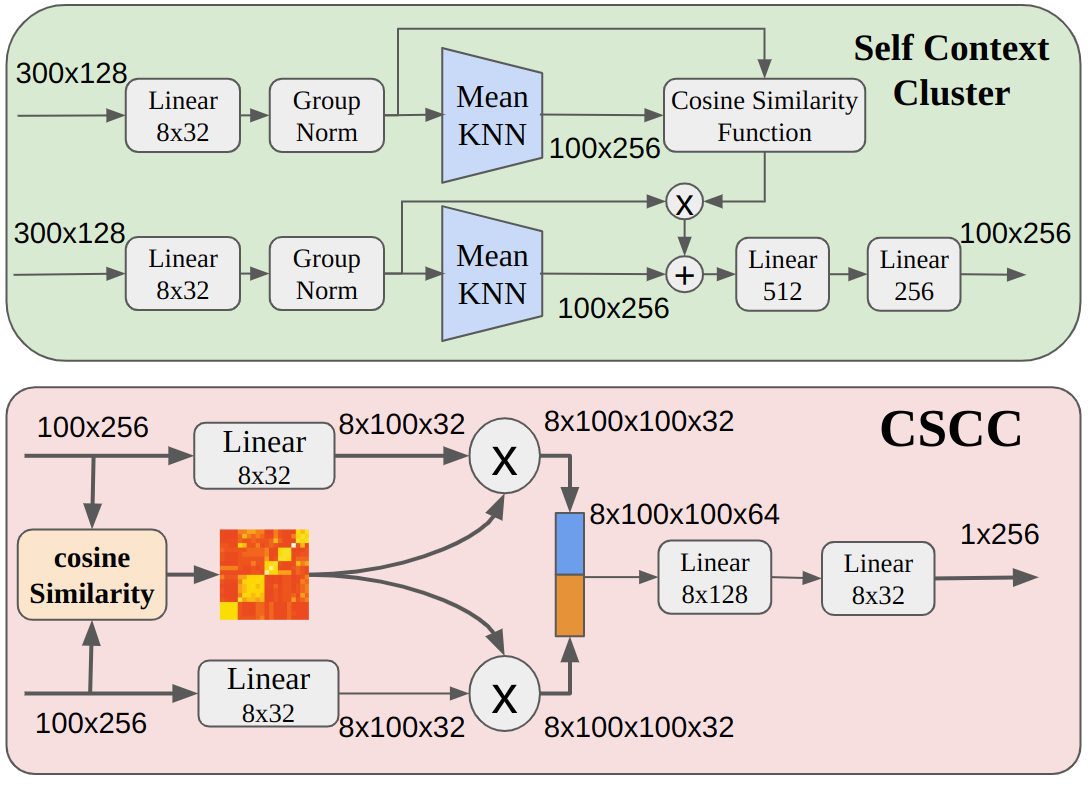}
    \caption{\textbf{Architecture Details: } The details of the architecture of SCC and CSCC.}
  \label{fig:details}
  \vspace{-1em}
\end{figure}

\begin{figure}
    \centering
    \includegraphics[width=\linewidth]{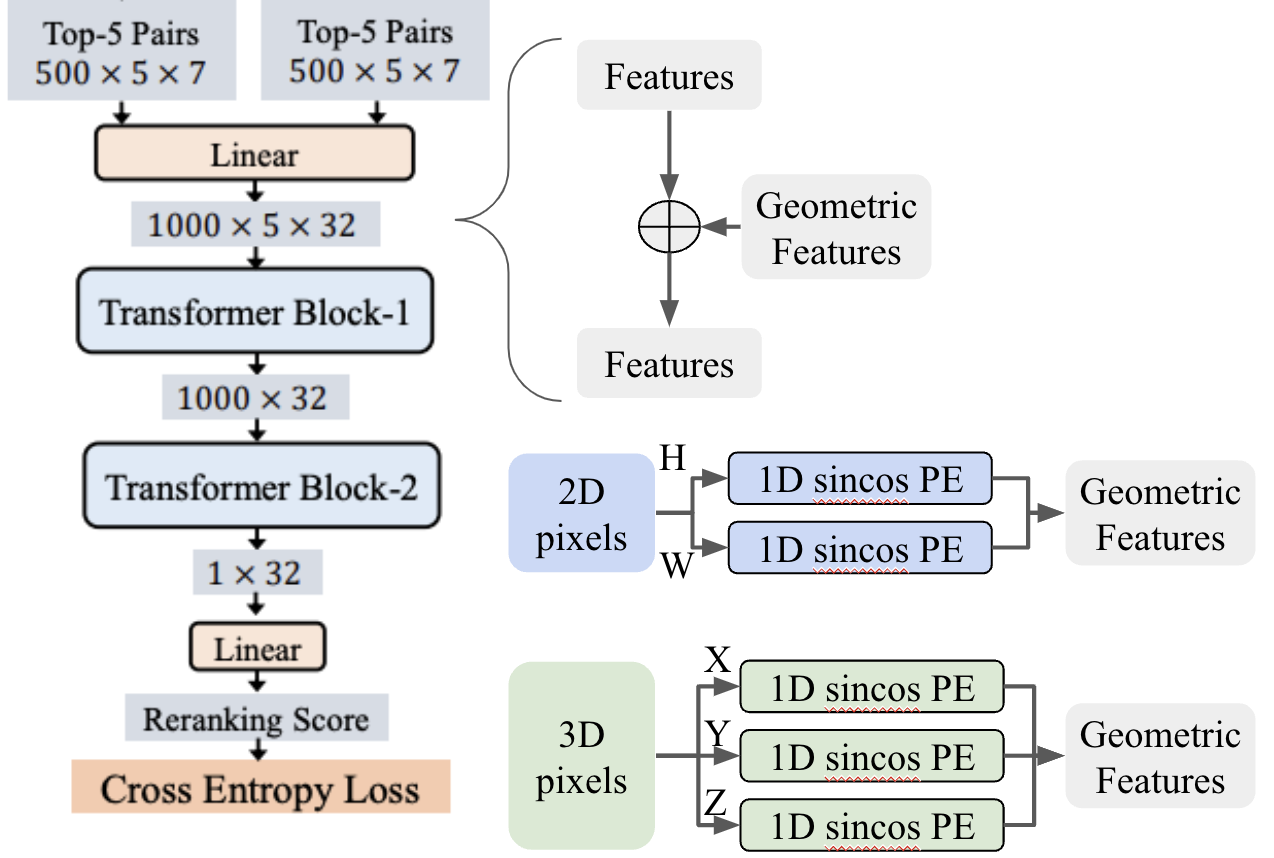}
    \caption{Our modified version adds one dimension to the geometric embedding process.}
    \label{fig:r2former}
    \vspace{-2em}
\end{figure}

\begin{figure}
  \centering
  \begin{tabular}{ |c|c|c|}
    \hline
   \textbf{Query Frame} & \textbf{CSCPR Recall@1} & \textbf{Ground truth}  \\ \hline

    \begin{minipage}{.3\linewidth} \includegraphics[width=\linewidth,height=0.75\linewidth]{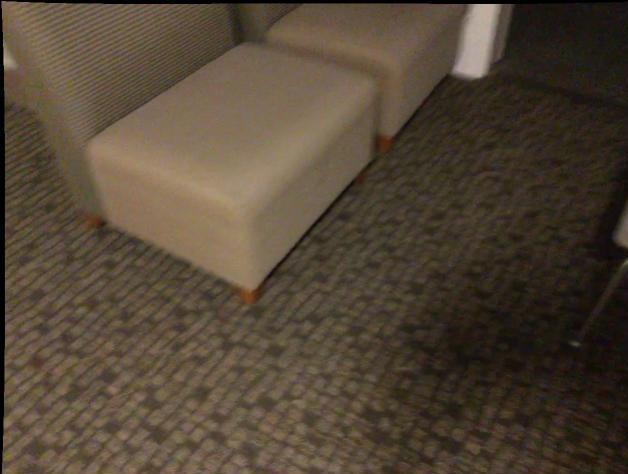} \end{minipage}
    &
    \begin{minipage}{.3\linewidth} \includegraphics[width=\linewidth,height=0.75\linewidth]{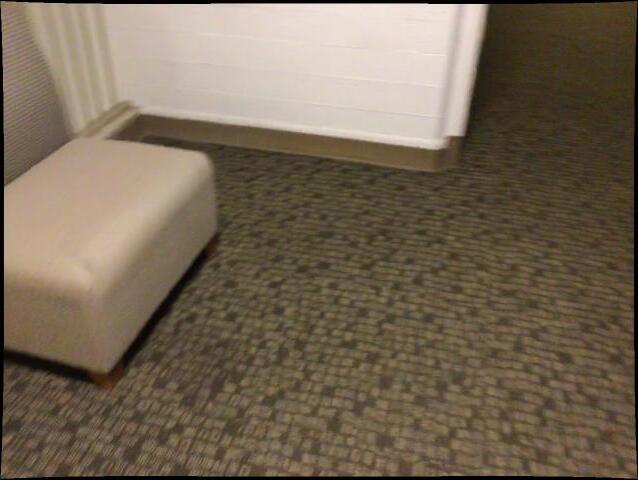} \end{minipage}
    &
    \begin{minipage}{.3\linewidth} \includegraphics[width=\linewidth,height=0.75\linewidth]{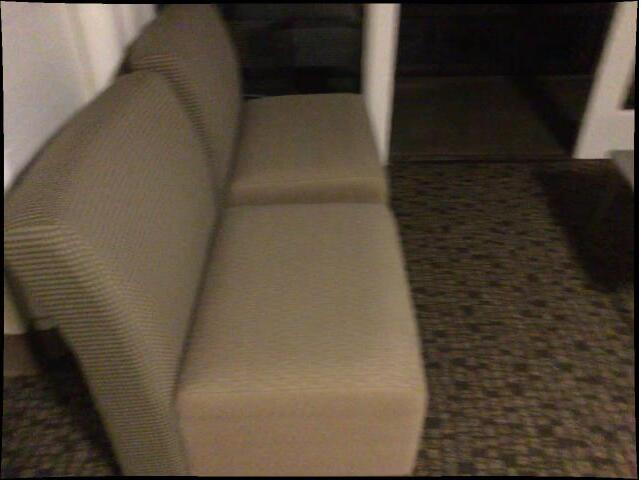} \end{minipage}
    \\ \hline

    \begin{minipage}{.3\linewidth} \includegraphics[width=\linewidth,height=0.75\linewidth]{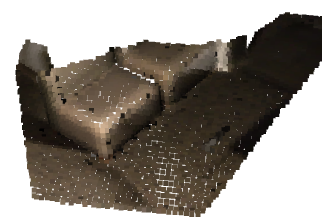} \end{minipage}
    &
    \begin{minipage}{.3\linewidth} \includegraphics[width=\linewidth,height=0.75\linewidth]{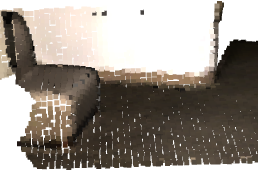} \end{minipage}
    &
    \begin{minipage}{.3\linewidth} \includegraphics[width=\linewidth,height=0.75\linewidth]{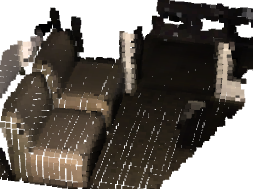} \end{minipage}
    \\ \hline
  \end{tabular}
  \caption{\textbf{Failure case: } Because the point cloud is not very clear, the top row of RGB images provides a clear view of the scenario. When there is not enough information in the point cloud to determine which one is correctly matched, CSCPR chooses the most similar structure and scale of the point cloud.}
  \label{fig:failure_cases}
  \vspace{-2em}
\end{figure}

\subsection{Dataset Differences} 
More of the dataset differences are shown in Fig.~\ref{fig:dataset_difference_all}.
\label{sec:dataset_differences}

\input{dataset_difference}

\subsection{Generated Datasets}

\begin{figure*}[!ht]
  \centering
  \begin{tabular}{ | c | c | c | c |}
    \hline
    \textbf{Scenes} & \textbf{RGB-D Points} & \textbf{Normals} & \textbf{Semantic} \\ \hline
    \multirow{2}{*}{
    \begin{minipage}{.25\linewidth}
      \centering
      \includegraphics[width=\linewidth,height=0.5\linewidth]{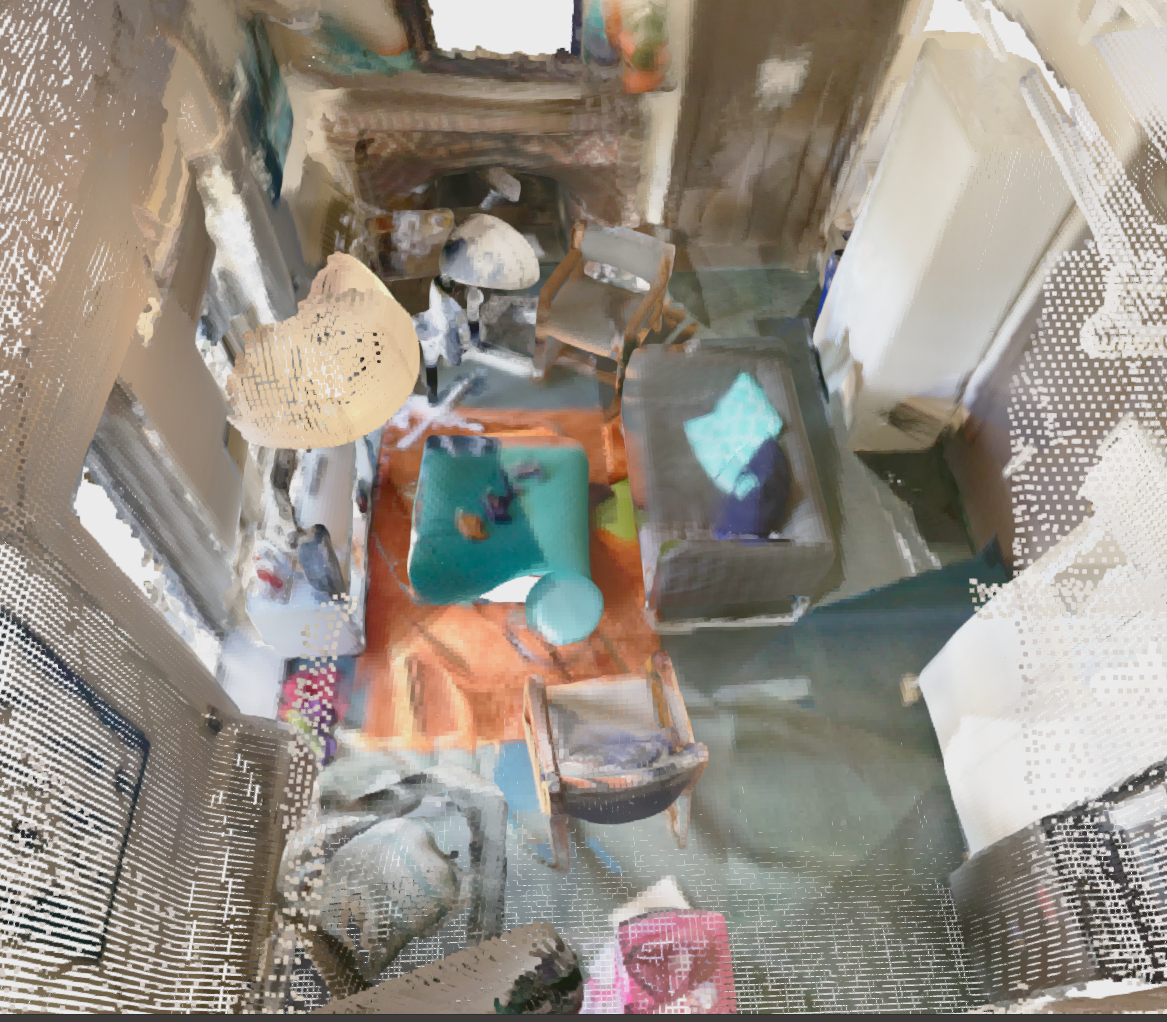}
    \end{minipage}
    } 
    & \begin{minipage}{.15\linewidth} \includegraphics[width=\linewidth,height=0.75\linewidth]{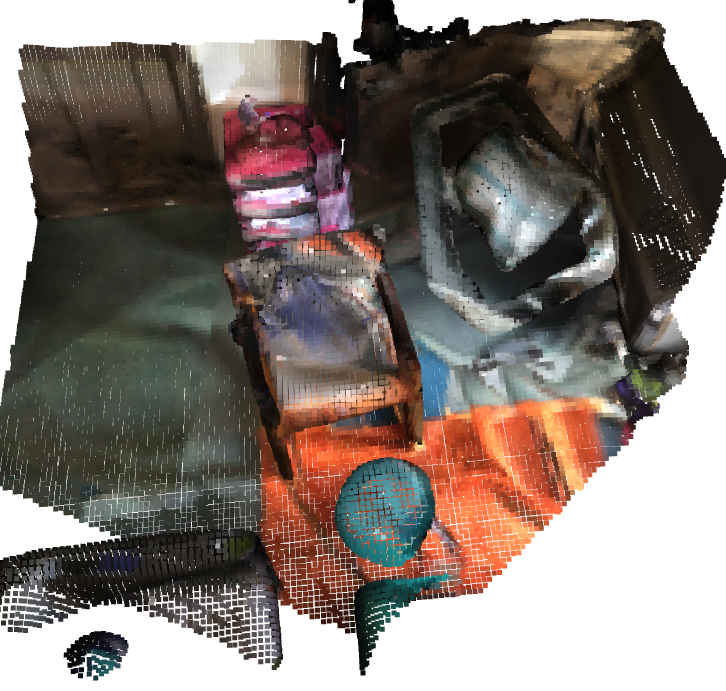} \end{minipage}
    & \begin{minipage}{.15\linewidth} \includegraphics[width=\linewidth,height=0.75\linewidth]{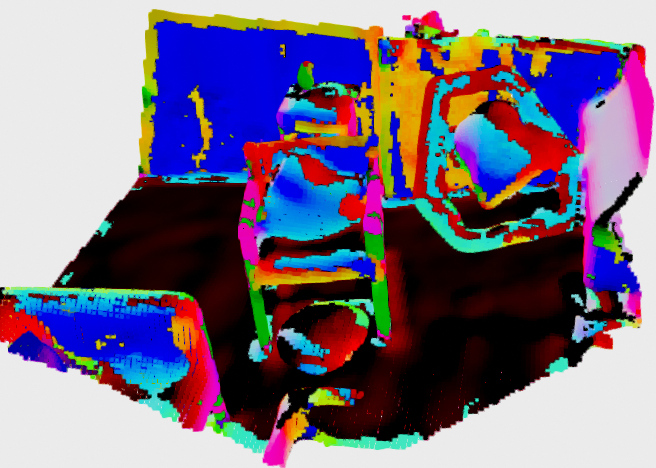} \end{minipage}
    & \begin{minipage}{.15\linewidth} \includegraphics[width=\linewidth,height=0.75\linewidth]{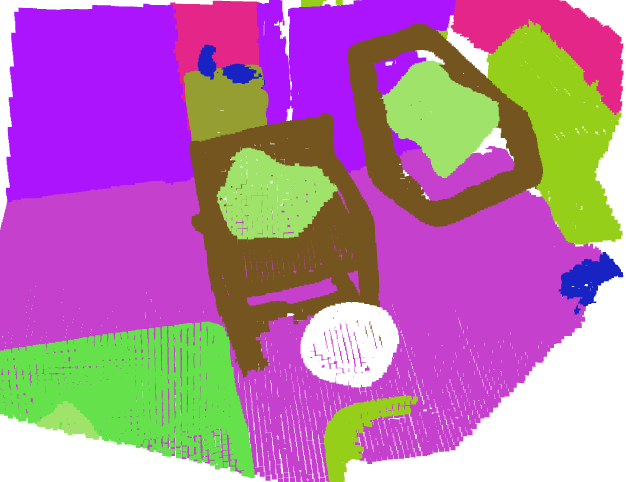} \end{minipage}\\ 
    \cline{2-4} 
    & \begin{minipage}{.15\linewidth} \includegraphics[width=\linewidth,height=0.75\linewidth]{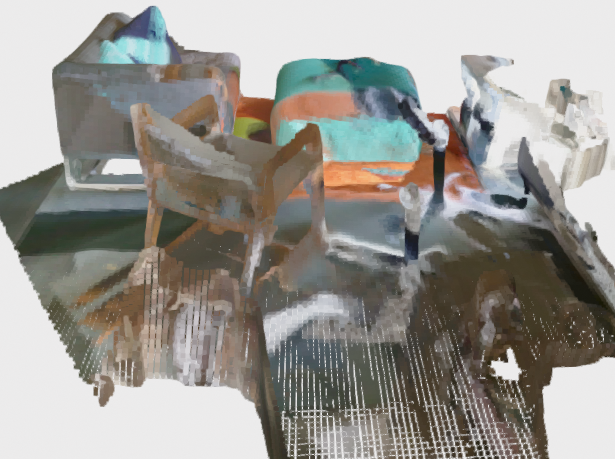} \end{minipage}
    & \begin{minipage}{.15\linewidth} \includegraphics[width=\linewidth,height=0.75\linewidth]{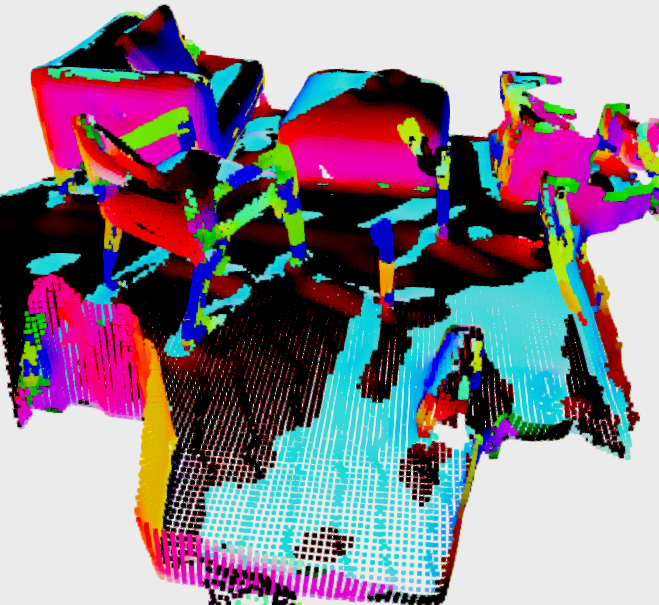} \end{minipage}
    & \begin{minipage}{.15\linewidth} \includegraphics[width=\linewidth,height=0.75\linewidth]{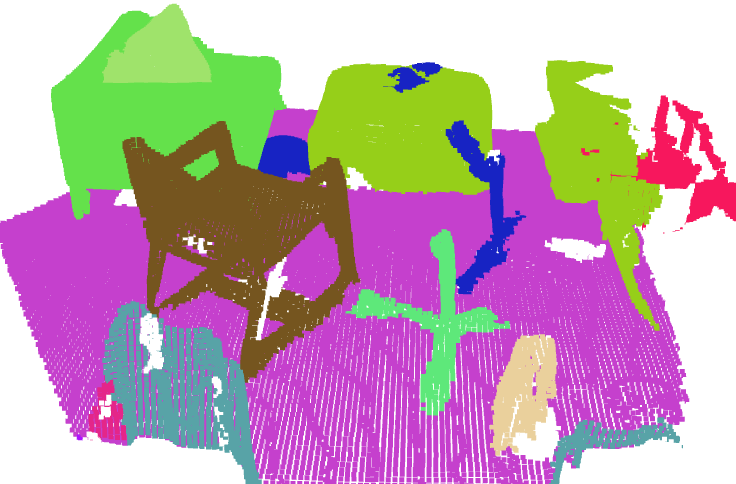} \end{minipage}\\ 
    \hline
    \multirow{2}{*}{
    \begin{minipage}{.25\linewidth}
      \centering
      \includegraphics[width=\linewidth,height=0.5\linewidth]{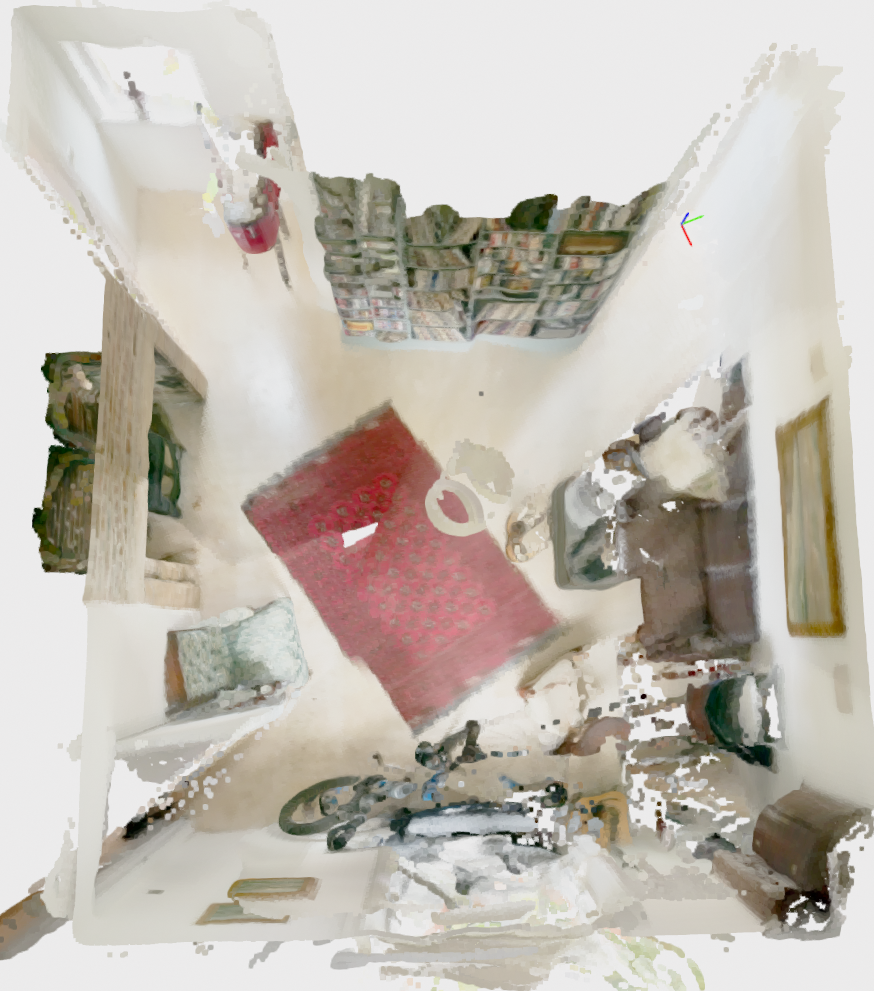}
    \end{minipage}
    } 
    & \begin{minipage}{.15\linewidth} \includegraphics[width=\linewidth,height=0.75\linewidth]{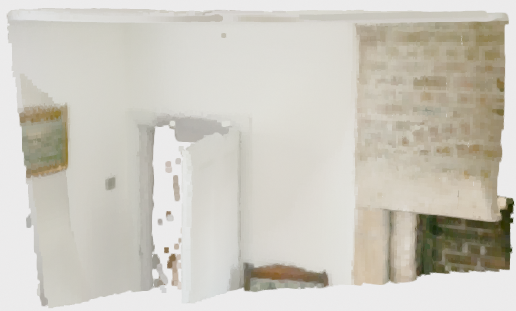} \end{minipage}
    & \begin{minipage}{.15\linewidth} \includegraphics[width=\linewidth,height=0.75\linewidth]{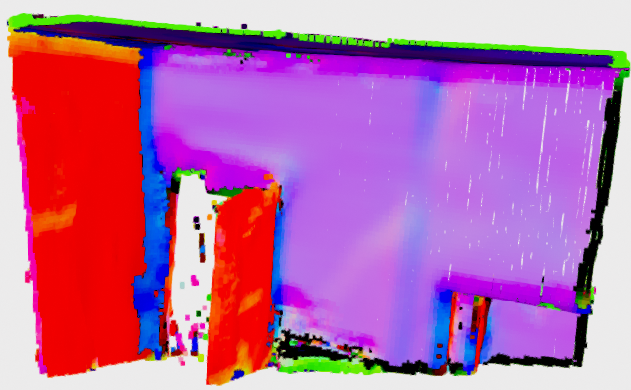} \end{minipage}
    & \begin{minipage}{.15\linewidth} \includegraphics[width=\linewidth,height=0.75\linewidth]{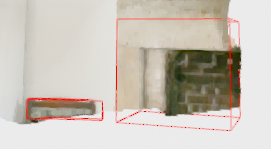} \end{minipage}\\ 
    \cline{2-4}
    & \begin{minipage}{.15\linewidth} \includegraphics[width=\linewidth,height=0.75\linewidth]{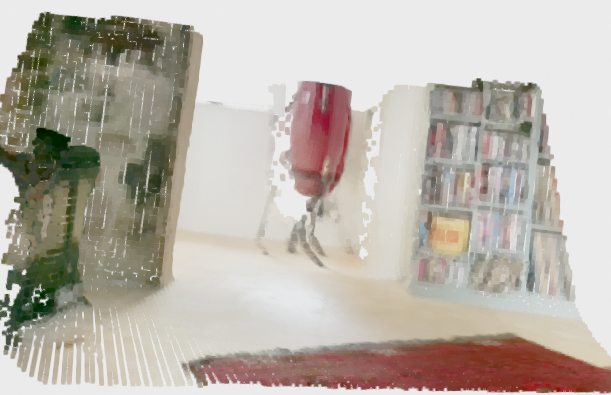} \end{minipage}
    & \begin{minipage}{.15\linewidth} \includegraphics[width=\linewidth,height=0.75\linewidth]{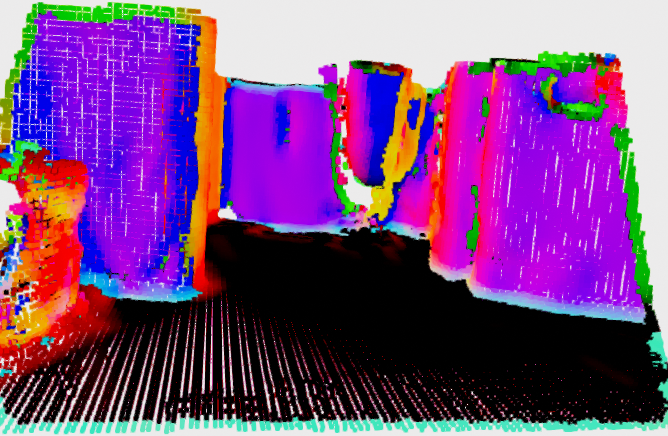} \end{minipage}
    & \begin{minipage}{.15\linewidth} \includegraphics[width=\linewidth,height=0.75\linewidth]{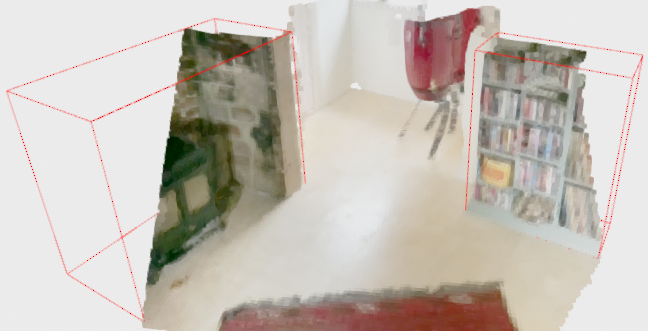} \end{minipage}\\ 
    \hline
  \end{tabular}
  \caption{\textbf{ScanNetIPR and ARKitIPR:} The top two rows show one scene of ScanNetIPR, where the normals are normal vectors that are perpendicular to the surface of the points. In ScanNetIPR, the semantic information is the segmentation of the points and ARKitIPR has the bounding boxes.} 
  \label{fig:dataset_diff}
\end{figure*}

As shown in Figure \ref{fig:dataset_diff}, the top row contains two point clouds in a scene of ScanNetIPR and the bottom row contains two point clouds in a scene of ARKitIPR. The normals in the figure represent the normal vectors of the points, which are shown as RGB images with the $\{n_x,n_y,n_z\}$ as $\{r,g,b\}$. The RGB-D point clouds are extracted as discussed in Section \ref{sec:data_generation}. The Semantic column is points segmentation in ScanNetIPR and is the object classification with bounding boxes in ARKitIPR.

\subsection{Architecture Details}
\label{sup:global_retrieval}
SCC and CSCC models are in Figure \ref{fig:details}. The input of the point features from the global retrieval has the shape of $300\times 128$, which has 300 points. Then the points are clustered into 100 centers with the dimension of 256. Finally, the centers are processed by the correlations between the two clouds of centers and then aggregated into a reranking score.

\subsection{Failure Analysis}
In Figure \ref{fig:failure_cases} the top row RGB images are to provide a clearer view of the scenario than the point cloud. As shown in Figure \ref{fig:failure_cases}, if there is not enough information to tell the difference between the best overlapped point cloud in the database and the query point cloud our model will struggle in detecting the best matched point cloud. Therefore, if more historic frames can be provided, more information can be used to correctly match the point clouds, and this is also potential future work to improve the performance of RGB-D indoor place recognition.

\subsection{Modified R2Former Reranking}
As shown in Figure \ref{fig:r2former}, the original R2former approach adds the geometric features to token features after the Linear layer. The geometric embedding method is the concatenation of two 1D sin and cos positional embeddings for $\set{H,W}$ 2D pixel positions. Our modification utilizes the same embedding approach, but to $\set{X,Y,Z}$, i.e. three dimensions.

%% file: dataset_difference.tex
\begin{figure*}[ht!]
  \centering
  \begin{tabular}{ | c | c | c | c | c |}
    \hline
    
    
        \multirow{2}{*}{\textbf{Query Frames}} & \multicolumn{2}{c|}{\textbf{FPF in ScanNetPR}} & \multicolumn{2}{c|}{\textbf{FPF in ScanNetIPR (Ours)}} \\
    \cline{2-5}
    & \textbf{Points} & \textbf{RGB} & \textbf{Points} & \textbf{RGB}
    \\ 
    \hline

    \begin{minipage}{.15\linewidth} \includegraphics[width=\linewidth,height=0.75\linewidth]{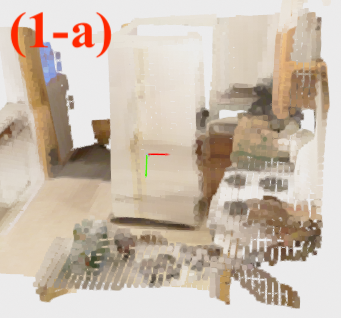} \end{minipage}
    &
    \begin{minipage}{.15\linewidth} \includegraphics[width=\linewidth,height=0.75\linewidth]{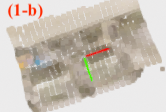} \end{minipage}
    &
    \begin{minipage}{.15\linewidth} \includegraphics[width=\linewidth,height=0.75\linewidth]{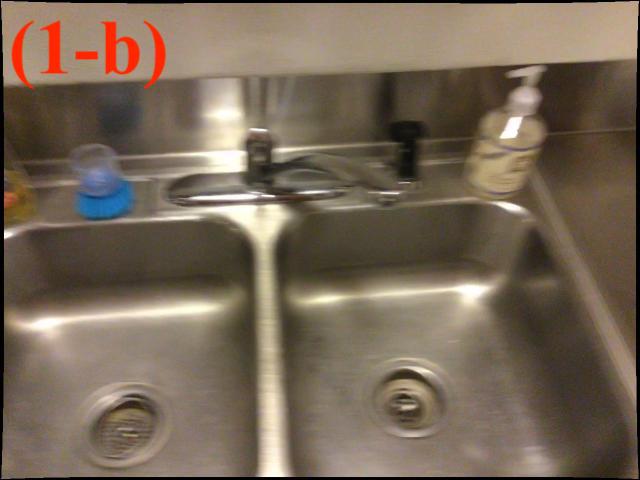} \end{minipage}
    &
    \begin{minipage}{.17\linewidth} \includegraphics[width=\linewidth,height=0.75\linewidth]{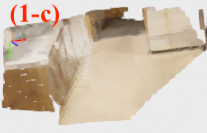} \end{minipage}
    &
    \begin{minipage}{.17\linewidth} \includegraphics[width=\linewidth,height=0.75\linewidth]{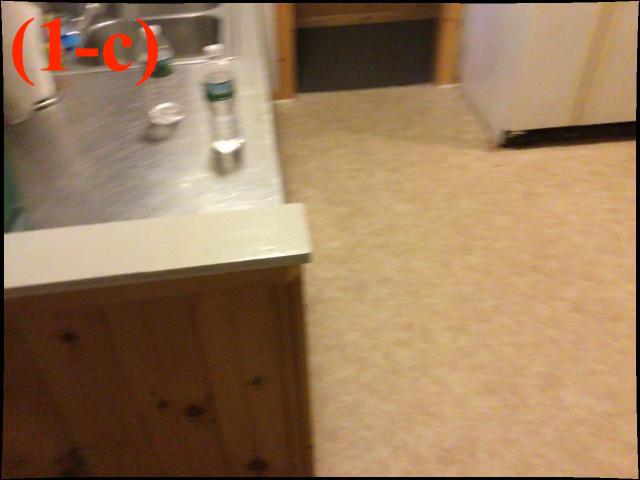} \end{minipage}
    \\ 
    \hline

    \begin{minipage}{.15\linewidth} \includegraphics[width=\linewidth,height=0.8\linewidth]{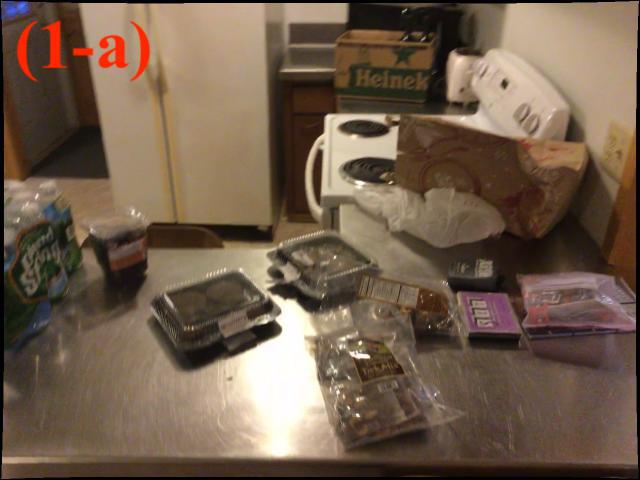} \end{minipage}
    &
    \multicolumn{2}{c|}{
    \begin{minipage}{.30\linewidth} \includegraphics[width=\linewidth,height=0.40\linewidth]{figs/dataset/1-14.png} \end{minipage}}
    &
    \multicolumn{2}{c|}{
    \begin{minipage}{.34\linewidth} \includegraphics[width=\linewidth,height=0.40\linewidth]{figs/dataset/1-18.png} \end{minipage}} 
    \\ 
    \hline\hline

    \begin{minipage}{.15\linewidth} \includegraphics[width=\linewidth,height=0.75\linewidth]{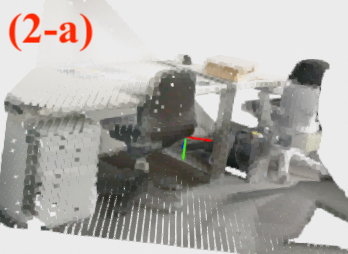} \end{minipage}
    &
    \begin{minipage}{.15\linewidth} \includegraphics[width=\linewidth,height=0.75\linewidth]{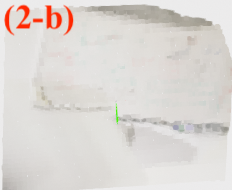} \end{minipage}
    &
    \begin{minipage}{.15\linewidth} \includegraphics[width=\linewidth,height=0.75\linewidth]{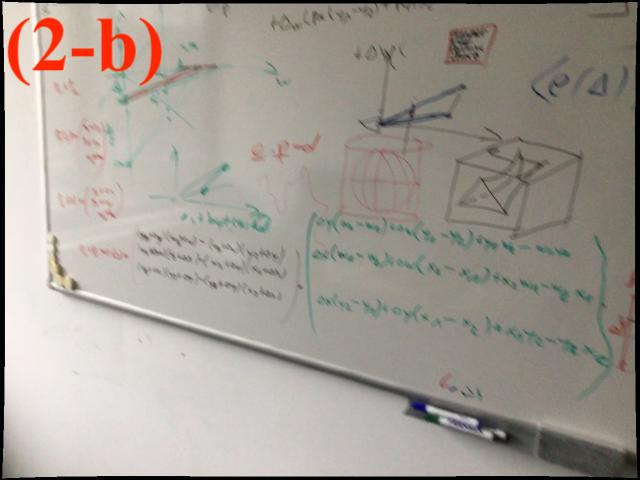} \end{minipage}
    &
    \begin{minipage}{.17\linewidth} \includegraphics[width=\linewidth,height=0.75\linewidth]{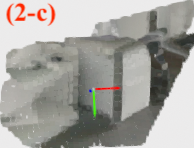} \end{minipage}
    &
    \begin{minipage}{.17\linewidth} \includegraphics[width=\linewidth,height=0.75\linewidth]{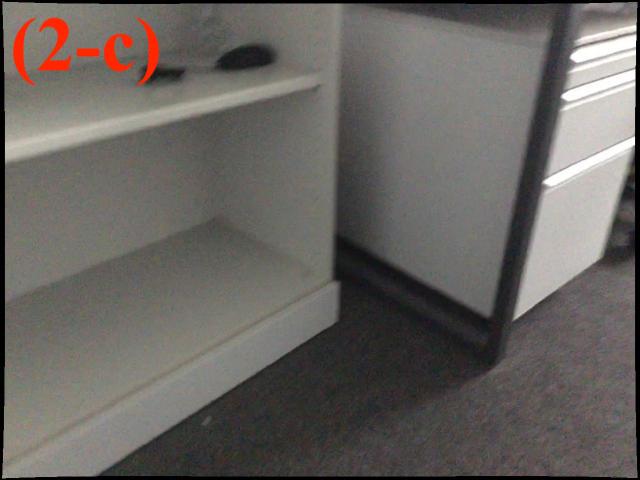} \end{minipage}
    \\ 
    \hline

    \begin{minipage}{.15\linewidth} \includegraphics[width=\linewidth,height=0.8\linewidth]{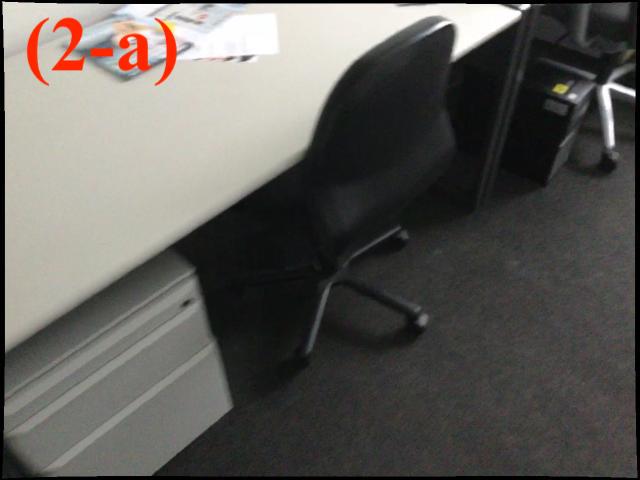} \end{minipage}
    &
    \multicolumn{2}{c|}{
    \begin{minipage}{.3\linewidth} \includegraphics[width=\linewidth, height=0.40\linewidth]{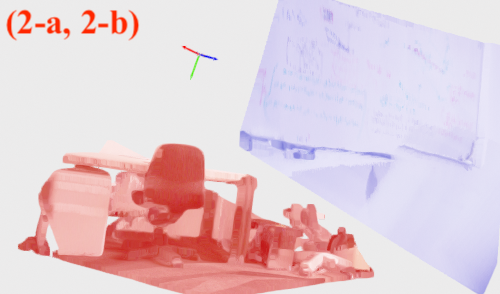} \end{minipage}}
    &
    \multicolumn{2}{c|}{
    \begin{minipage}{.32\linewidth} \includegraphics[width=\linewidth, height=0.40\linewidth]{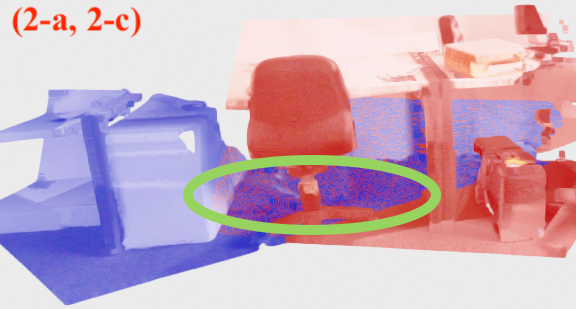} \end{minipage}} 
    \\ 
    \hline

  \end{tabular}
  \caption{\textbf{Furthest Positive Frame (FPF) of ScanNetPR vs. ScanNetIPR (ours):} FPF depicts the matched frame that overlaps least with the query in both datasets. For the red letters: $\set{1, 2}$ represent two cases and $\set{a,b,c}$ are query, ScanNetPR, and ScanNetIPR frames, respectively. In the first column, the 1st and 3rd rows show point clouds, and the 2nd and 4th rows show the corresponding RGB images for a clearer view. In later columns, the 2nd and 4th rows show the query (red) and FPF (blue) frames in the global coordinate and their overlapped areas (green circles). In the ScanNetPR dataset, using the center distance as the criterion to determine matched frames leads to erroneous classification. In ScanNetIPR, the overlap is the only criterion for matching; thus it is more accurate for training and evaluating place recognition tasks.} 
  \label{fig:dataset_difference_all}
\end{figure*}

%% file: root.bbl
\begin{thebibliography}{10}
\providecommand{\url}[1]{#1}
\csname url@samestyle\endcsname
\providecommand{\newblock}{\relax}
\providecommand{\bibinfo}[2]{#2}
\providecommand{\BIBentrySTDinterwordspacing}{\spaceskip=0pt\relax}
\providecommand{\BIBentryALTinterwordstretchfactor}{4}
\providecommand{\BIBentryALTinterwordspacing}{\spaceskip=\fontdimen2\font plus
\BIBentryALTinterwordstretchfactor\fontdimen3\font minus \fontdimen4\font\relax}
\providecommand{\BIBforeignlanguage}[2]{{%
\expandafter\ifx\csname l@#1\endcsname\relax
\typeout{** WARNING: IEEEtran.bst: No hyphenation pattern has been}%
\typeout{** loaded for the language `#1'. Using the pattern for}%
\typeout{** the default language instead.}%
\else
\language=\csname l@#1\endcsname
\fi
#2}}
\providecommand{\BIBdecl}{\relax}
\BIBdecl

\bibitem{liang2024poco}
J.~Liang, Z.~Deng, Z.~Zhou, O.~Ghasemalizadeh, D.~Manocha, M.~Sun, C.-H. Kuo, and S.~Arnie, ``Poco: Point context cluster for rgbd indoor place recognition,'' in \emph{2024 IEEE/RSJ International Conference on Intelligent Robots and Systems (IROS)}.\hskip 1em plus 0.5em minus 0.4em\relax IEEE, 2024.

\bibitem{cocs}
\BIBentryALTinterwordspacing
X.~Ma, Y.~Zhou, H.~Wang, C.~Qin, B.~Sun, C.~Liu, and Y.~Fu, ``Image as set of points,'' in \emph{The Eleventh International Conference on Learning Representations}, 2023. [Online]. Available: \url{https://openreview.net/forum?id=awnvqZja69}
\BIBentrySTDinterwordspacing

\bibitem{ijcai2021p603}
\BIBentryALTinterwordspacing
S.~Garg, T.~Fischer, and M.~Milford, ``Where is your place, visual place recognition?'' in \emph{Proceedings of the Thirtieth International Joint Conference on Artificial Intelligence, {IJCAI-21}}, Z.-H. Zhou, Ed.\hskip 1em plus 0.5em minus 0.4em\relax International Joint Conferences on Artificial Intelligence Organization, 8 2021, pp. 4416--4425, survey Track. [Online]. Available: \url{https://doi.org/10.24963/ijcai.2021/603}
\BIBentrySTDinterwordspacing

\bibitem{lowry2015visual}
S.~Lowry, N.~S{\"u}nderhauf, P.~Newman, J.~J. Leonard, D.~Cox, P.~Corke, and M.~J. Milford, ``Visual place recognition: A survey,'' \emph{ieee transactions on robotics}, vol.~32, no.~1, pp. 1--19, 2015.

\bibitem{liu2023survey}
Y.~Liu, Y.~Zhang, Y.~Wang, F.~Hou, J.~Yuan, J.~Tian, Y.~Zhang, Z.~Shi, J.~Fan, and Z.~He, ``A survey of visual transformers,'' \emph{IEEE Transactions on Neural Networks and Learning Systems}, 2023.

\bibitem{yurtsever2020survey}
E.~Yurtsever, J.~Lambert, A.~Carballo, and K.~Takeda, ``A survey of autonomous driving: Common practices and emerging technologies,'' \emph{IEEE access}, vol.~8, pp. 58\,443--58\,469, 2020.

\bibitem{kornilova2023dominating}
A.~Kornilova, I.~Moskalenko, T.~Pushkin, F.~Tojiboev, R.~Tariverdizadeh, and G.~Ferrer, ``Dominating set database selection for visual place recognition,'' \emph{arXiv preprint arXiv:2303.05123}, 2023.

\bibitem{middelberg2014scalable}
S.~Middelberg, T.~Sattler, O.~Untzelmann, and L.~Kobbelt, ``Scalable 6-dof localization on mobile devices,'' in \emph{Computer Vision--ECCV 2014: 13th European Conference, Zurich, Switzerland, September 6-12, 2014, Proceedings, Part II 13}.\hskip 1em plus 0.5em minus 0.4em\relax Springer, 2014, pp. 268--283.

\bibitem{mirowski2018learning}
P.~Mirowski, M.~Grimes, M.~Malinowski, K.~M. Hermann, K.~Anderson, D.~Teplyashin, K.~Simonyan, A.~Zisserman, R.~Hadsell \emph{et~al.}, ``Learning to navigate in cities without a map,'' \emph{Advances in neural information processing systems}, vol.~31, 2018.

\bibitem{bresson2017simultaneous}
G.~Bresson, Z.~Alsayed, L.~Yu, and S.~Glaser, ``Simultaneous localization and mapping: A survey of current trends in autonomous driving,'' \emph{IEEE Transactions on Intelligent Vehicles}, vol.~2, no.~3, pp. 194--220, 2017.

\bibitem{cgis}
Y.~Ming, X.~Yang, G.~Zhang, and A.~Calway, ``Cgis-net: Aggregating colour, geometry and implicit semantic features for indoor place recognition,'' in \emph{2022 IEEE/RSJ International Conference on Intelligent Robots and Systems (IROS)}.\hskip 1em plus 0.5em minus 0.4em\relax IEEE, 2022, pp. 6991--6997.

\bibitem{du2020dh3d}
J.~Du, R.~Wang, and D.~Cremers, ``Dh3d: Deep hierarchical 3d descriptors for robust large-scale 6dof relocalization,'' in \emph{Computer Vision--ECCV 2020: 16th European Conference, Glasgow, UK, August 23--28, 2020, Proceedings, Part IV 16}.\hskip 1em plus 0.5em minus 0.4em\relax Springer, 2020, pp. 744--762.

\bibitem{zhu2023r2former}
S.~Zhu, L.~Yang, C.~Chen, M.~Shah, X.~Shen, and H.~Wang, ``R2former: Unified retrieval and reranking transformer for place recognition,'' in \emph{Proceedings of the IEEE/CVF Conference on Computer Vision and Pattern Recognition}, 2023, pp. 19\,370--19\,380.

\bibitem{vaswani2017attention}
A.~Vaswani, N.~Shazeer, N.~Parmar, J.~Uszkoreit, L.~Jones, A.~N. Gomez, L.~Kaiser, and I.~Polosukhin, ``Attention is all you need,'' \emph{Advances in neural information processing systems}, vol.~30, 2017.

\bibitem{uy2018pointnetvlad}
M.~A. Uy and G.~H. Lee, ``Pointnetvlad: Deep point cloud based retrieval for large-scale place recognition,'' in \emph{Proceedings of the IEEE conference on computer vision and pattern recognition}, 2018, pp. 4470--4479.

\bibitem{arandjelovic2016netvlad}
R.~Arandjelovic, P.~Gronat, A.~Torii, T.~Pajdla, and J.~Sivic, ``Netvlad: Cnn architecture for weakly supervised place recognition,'' in \emph{Proceedings of the IEEE conference on computer vision and pattern recognition}, 2016, pp. 5297--5307.

\bibitem{hausler2021patch}
S.~Hausler, S.~Garg, M.~Xu, M.~Milford, and T.~Fischer, ``Patch-netvlad: Multi-scale fusion of locally-global descriptors for place recognition,'' in \emph{Proceedings of the IEEE/CVF Conference on Computer Vision and Pattern Recognition}, 2021, pp. 14\,141--14\,152.

\bibitem{vidanapathirana2023spectral}
K.~Vidanapathirana, P.~Moghadam, S.~Sridharan, and C.~Fookes, ``Spectral geometric verification: Re-ranking point cloud retrieval for metric localization,'' \emph{IEEE Robotics and Automation Letters}, vol.~8, no.~5, pp. 2494--2501, 2023.

\bibitem{ransac}
M.~A. Fischler and R.~C. Bolles, ``Random sample consensus: a paradigm for model fitting with applications to image analysis and automated cartography,'' \emph{Communications of the ACM}, vol.~24, no.~6, pp. 381--395, 1981.

\bibitem{tan2021instance}
F.~Tan, J.~Yuan, and V.~Ordonez, ``Instance-level image retrieval using reranking transformers,'' in \emph{proceedings of the IEEE/CVF international conference on computer vision}, 2021, pp. 12\,105--12\,115.

\bibitem{lee2022correlation}
S.~Lee, H.~Seong, S.~Lee, and E.~Kim, ``Correlation verification for image retrieval,'' in \emph{Proceedings of the IEEE/CVF Conference on Computer Vision and Pattern Recognition}, 2022, pp. 5374--5384.

\bibitem{dai2017scannet}
A.~Dai, A.~X. Chang, M.~Savva, M.~Halber, T.~Funkhouser, and M.~Nie{\ss}ner, ``Scannet: Richly-annotated 3d reconstructions of indoor scenes,'' in \emph{Proc. Computer Vision and Pattern Recognition (CVPR), IEEE}, 2017.

\bibitem{dehghan2021arkitscenes}
\BIBentryALTinterwordspacing
G.~Baruch, Z.~Chen, A.~Dehghan, T.~Dimry, Y.~Feigin, P.~Fu, T.~Gebauer, B.~Joffe, D.~Kurz, A.~Schwartz, and E.~Shulman, ``{ARK}itscenes - a diverse real-world dataset for 3d indoor scene understanding using mobile {RGB}-d data,'' in \emph{Thirty-fifth Conference on Neural Information Processing Systems Datasets and Benchmarks Track (Round 1)}, 2021. [Online]. Available: \url{https://openreview.net/forum?id=tjZjv_qh_CE}
\BIBentrySTDinterwordspacing

\bibitem{yudin2022hpointloc}
D.~Yudin, Y.~Solomentsev, R.~Musaev, A.~Staroverov, and A.~I. Panov, ``Hpointloc: Point-based indoor place recognition using synthetic rgb-d images,'' in \emph{International Conference on Neural Information Processing}.\hskip 1em plus 0.5em minus 0.4em\relax Springer, 2022, pp. 471--484.

\bibitem{david2004distinctive}
L.~David, ``Distinctive image features from scale-invariant keypoints,'' \emph{International journal of computer vision}, vol.~60, pp. 91--110, 2004.

\bibitem{cummins2008fab}
M.~Cummins and P.~Newman, ``Fab-map: Probabilistic localization and mapping in the space of appearance,'' \emph{The International journal of robotics research}, vol.~27, no.~6, pp. 647--665, 2008.

\bibitem{galvez2012bags}
D.~G{\'a}lvez-L{\'o}pez and J.~D. Tardos, ``Bags of binary words for fast place recognition in image sequences,'' \emph{IEEE Transactions on Robotics}, vol.~28, no.~5, pp. 1188--1197, 2012.

\bibitem{gordo2017end}
A.~Gordo, J.~Almazan, J.~Revaud, and D.~Larlus, ``End-to-end learning of deep visual representations for image retrieval,'' \emph{International Journal of Computer Vision}, vol. 124, no.~2, pp. 237--254, 2017.

\bibitem{berton2022deep}
G.~Berton, R.~Mereu, G.~Trivigno, C.~Masone, G.~Csurka, T.~Sattler, and B.~Caputo, ``Deep visual geo-localization benchmark,'' in \emph{Proceedings of the IEEE/CVF Conference on Computer Vision and Pattern Recognition}, 2022, pp. 5396--5407.

\bibitem{touvron2021training}
H.~Touvron, M.~Cord, M.~Douze, F.~Massa, A.~Sablayrolles, and H.~J{\'e}gou, ``Training data-efficient image transformers \& distillation through attention,'' in \emph{International conference on machine learning}.\hskip 1em plus 0.5em minus 0.4em\relax PMLR, 2021, pp. 10\,347--10\,357.

\bibitem{mur2017orb}
R.~Mur-Artal and J.~D. Tard{\'o}s, ``Orb-slam2: An open-source slam system for monocular, stereo, and rgb-d cameras,'' \emph{IEEE transactions on robotics}, vol.~33, no.~5, pp. 1255--1262, 2017.

\bibitem{komorowski2021minkloc3d}
J.~Komorowski, ``Minkloc3d: Point cloud based large-scale place recognition,'' in \emph{Proceedings of the IEEE/CVF Winter Conference on Applications of Computer Vision}, 2021, pp. 1790--1799.

\bibitem{thomas2019kpconv}
H.~Thomas, C.~R. Qi, J.-E. Deschaud, B.~Marcotegui, F.~Goulette, and L.~J. Guibas, ``Kpconv: Flexible and deformable convolution for point clouds,'' in \emph{Proceedings of the IEEE/CVF international conference on computer vision}, 2019, pp. 6411--6420.

\bibitem{zhang2022rank}
W.~Zhang, H.~Zhou, Z.~Dong, Q.~Yan, and C.~Xiao, ``Rank-pointretrieval: Reranking point cloud retrieval via a visually consistent registration evaluation,'' \emph{IEEE Transactions on Visualization and Computer Graphics}, 2022.

\bibitem{Barath_2020_CVPR}
D.~Barath, J.~Noskova, M.~Ivashechkin, and J.~Matas, ``Magsac++, a fast, reliable and accurate robust estimator,'' in \emph{Proceedings of the IEEE/CVF Conference on Computer Vision and Pattern Recognition (CVPR)}, June 2020.

\bibitem{urr}
B.~et~al., ``Unsupervisedr\&r: Unsupervised point cloud registration via differentiable rendering,'' 2021.

\bibitem{mei2023unsupervised}
G.~Mei, H.~Tang, X.~Huang, W.~Wang, J.~Liu, J.~Zhang, L.~Van~Gool, and Q.~Wu, ``Unsupervised deep probabilistic approach for partial point cloud registration,'' in \emph{Proceedings of the IEEE/CVF Conference on Computer Vision and Pattern Recognition}, 2023, pp. 13\,611--13\,620.

\bibitem{hatem2023point}
A.~Hatem, Y.~Qian, and Y.~Wang, ``Point-tta: Test-time adaptation for point cloud registration using multitask meta-auxiliary learning,'' in \emph{Proceedings of the IEEE/CVF International Conference on Computer Vision}, 2023, pp. 16\,494--16\,504.

\bibitem{li20232d3d}
M.~Li, Z.~Qin, Z.~Gao, R.~Yi, C.~Zhu, Y.~Guo, and K.~Xu, ``2d3d-matr: 2d-3d matching transformer for detection-free registration between images and point clouds,'' in \emph{Proceedings of the IEEE/CVF International Conference on Computer Vision}, 2023, pp. 14\,128--14\,138.

\bibitem{sarlin2020superglue}
P.-E. Sarlin, D.~DeTone, T.~Malisiewicz, and A.~Rabinovich, ``Superglue: Learning feature matching with graph neural networks,'' in \emph{Proceedings of the IEEE/CVF conference on computer vision and pattern recognition}, 2020, pp. 4938--4947.

\bibitem{dosovitskiy2020image}
A.~Dosovitskiy, L.~Beyer, A.~Kolesnikov, D.~Weissenborn, X.~Zhai, T.~Unterthiner, M.~Dehghani, M.~Minderer, G.~Heigold, S.~Gelly \emph{et~al.}, ``An image is worth 16x16 words: Transformers for image recognition at scale,'' \emph{arXiv preprint arXiv:2010.11929}, 2020.

\bibitem{loshchilov2016sgdr}
I.~Loshchilov and F.~Hutter, ``Sgdr: Stochastic gradient descent with warm restarts,'' \emph{arXiv preprint arXiv:1608.03983}, 2016.

\bibitem{kingma2014adam}
D.~P. Kingma and J.~Ba, ``Adam: A method for stochastic optimization,'' \emph{arXiv preprint arXiv:1412.6980}, 2014.

\bibitem{robinson2020contrastive}
J.~Robinson, C.-Y. Chuang, S.~Sra, and S.~Jegelka, ``Contrastive learning with hard negative samples,'' \emph{arXiv preprint arXiv:2010.04592}, 2020.

\bibitem{arun1987least}
K.~S. Arun, T.~S. Huang, and S.~D. Blostein, ``Least-squares fitting of two 3-d point sets,'' \emph{IEEE Transactions on pattern analysis and machine intelligence}, no.~5, pp. 698--700, 1987.

\bibitem{yang2020teaser}
H.~Yang, J.~Shi, and L.~Carlone, ``Teaser: Fast and certifiable point cloud registration,'' \emph{IEEE Transactions on Robotics}, vol.~37, no.~2, pp. 314--333, 2020.

\bibitem{barath2020magsac}
D.~Barath, J.~Noskova, M.~Ivashechkin, and J.~Matas, ``Magsac++, a fast, reliable and accurate robust estimator,'' in \emph{Proceedings of the IEEE/CVF conference on computer vision and pattern recognition}, 2020, pp. 1304--1312.

\bibitem{lu2024towards}
F.~Lu, L.~Zhang, X.~Lan, S.~Dong, Y.~Wang, and C.~Yuan, ``Towards seamless adaptation of pre-trained models for visual place recognition,'' \emph{arXiv preprint arXiv:2402.14505}, 2024.

\bibitem{pointmbf}
\BIBentryALTinterwordspacing
M.~Yuan, K.~Fu, Z.~Li, Y.~Meng, and M.~Wang, ``Pointmbf: A multi-scale bidirectional fusion network for unsupervised rgb-d point cloud registration,'' in \emph{2023 IEEE/CVF International Conference on Computer Vision (ICCV)}.\hskip 1em plus 0.5em minus 0.4em\relax Los Alamitos, CA, USA: IEEE Computer Society, oct 2023, pp. 17\,648--17\,659. [Online]. Available: \url{https://doi.ieeecomputersociety.org/10.1109/ICCV51070.2023.01622}
\BIBentrySTDinterwordspacing

\bibitem{sivic2008efficient}
J.~Sivic and A.~Zisserman, ``Efficient visual search of videos cast as text retrieval,'' \emph{IEEE transactions on pattern analysis and machine intelligence}, vol.~31, no.~4, pp. 591--606, 2008.

\bibitem{ming2024aegis}
Y.~Ming, J.~Ma, X.~Yang, W.~Dai, Y.~Peng, and W.~Kong, ``Aegis-net: Attention-guided multi-level feature aggregation for indoor place recognition,'' in \emph{ICASSP 2024-2024 IEEE International Conference on Acoustics, Speech and Signal Processing (ICASSP)}.\hskip 1em plus 0.5em minus 0.4em\relax IEEE, 2024, pp. 4030--4034.

\end{thebibliography}
